\documentclass{article}
\usepackage{iclr2027_conference,times}
\usepackage{iftex}
\ifPDFTeX
  \usepackage[T1]{fontenc}
\else
  \usepackage{fontspec}
\fi

\usepackage{hyperref}
\usepackage{url}
\usepackage{booktabs}
\usepackage{graphicx}
\usepackage{amsmath}
\usepackage{amssymb}
\usepackage[capitalise,nameinlink]{cleveref}
\crefname{section}{Sec.}{Secs.}
\Crefname{section}{Sec.}{Secs.}
\crefname{appendix}{App.}{Apps.}
\Crefname{appendix}{App.}{Apps.}
\newcommand{\TV}{\mathrm{TV}}
\newcommand{\E}{\mathbb{E}}
\newcommand{\Iset}{\mathcal{I}}
\newcommand{\Tz}{T^{(0)}}
\newcommand{\To}{T^{(1)}}

\newcommand{\papertitle}{Beyond Parallel Blindness: Information Floors and Model Gaps in Block Drafting}
\title{\papertitle}

\author{Xinwei Qiang \quad Xiang Fang \quad Chang Chen \quad Zaifeng Pan \quad Yue Guan \quad Yufei Ding\\
University of California San Diego\\
\texttt{\{x2qiang,x8fang,chc278,zapan,y9guan,yufeiding\}@ucsd.edu}}

\iclrfinalcopy
\hypersetup{
  hidelinks,
  pdftitle={Beyond Parallel Blindness: Information Floors and Model Gaps in Block Drafting},
  pdfauthor={Xinwei Qiang, Xiang Fang, Chang Chen, Zaifeng Pan, Yue Guan, Yufei Ding}
}

\begin{document}
\maketitle
\pagestyle{plain}
\thispagestyle{plain}

\begin{abstract}
Block drafters propose several tokens in one forward pass, before earlier target tokens are
realised. Their rejection mixes two losses: missing within-block path information and imperfect
modelling of observable information. Accepted length cannot distinguish them. We separate the two
with an \emph{information floor}, the minimum expected rejection at a specified conditioning order;
rejection above this floor is the \emph{model gap}. Measurements across four domains and four open-weight targets, plus floor estimates on a frontier API target, yield three findings.
First, the all-parallel floor reaches $0.286$ at the final slot on Qwen3-4B, limiting even the best proposal to
$71\%$ per-slot acceptance. Second, one realised token removes $86$--$100\%$ of this floor, a
locality also recovered by an independent mutual-information analysis. Third, current drafters
remain far above their floors: the final-slot model gap accounts for $43$--$64\%$ of DFlash
rejection and $85$--$92\%$ of DSpark's oracle-conditioned rejection. Guided by this separation of short-range conditioning from proposal quality, we replace
a context-independent predecessor correction with a prefix-attention head that reads committed
context conditional on the predecessor. With supporting backbone components, the resulting
Qwen3-4B drafter improves mean serving accepted length by $3.93\%$ over the released DSpark
checkpoint across nine tasks, without increasing the conditioning order.
Code and data: \url{https://github.com/tie-pilot-qxw/specfloor}; drafter checkpoints:
\url{https://huggingface.co/TIE-Pilot/dspark-attnconv-block7-qwen3-4b}.
\end{abstract}

\section{Introduction}

Speculative decoding \citep{leviathan2023fast,chen2023acceleratinglargelanguagemodel} accelerates autoregressive
generation by using a lightweight drafter to propose several future tokens and a target model to
verify them under a rule that preserves the target distribution exactly. Its speedup depends on how
many proposed tokens the target accepts. Recent \emph{block} drafters produce proposals for an entire
block in one forward pass \citep{pmlr-v235-cai24b,chen2026dflashblockdiffusionflash,cheng2026dsparkconfidencescheduledspeculativedecoding}, making drafting highly
parallel. In an all-parallel proposal, position $k$ faces a basic information constraint:
its distribution must be fixed without seeing the tokens at earlier block positions.

Current block drafters deliver state-of-the-art accepted lengths, making it natural to ask whether
all-parallel drafting is already close to its intrinsic limit. Accepted length alone cannot answer
this question: it records the rejection of a particular drafter but provides no reference for the
smallest rejection achievable under the same information constraint. The remaining loss may
therefore have two sources: the absence of earlier target realisations when the block is drafted, and
the drafter's imperfect use of the information it can observe. Separating them is necessary to
determine where further progress can come from.

The two losses point to different routes for improving block drafting. Information loss calls for
changing what a draft position can observe, for example by conditioning it on earlier realised
tokens. Loss above the information limit concerns how effectively the drafter operates with the
information already available, through its model, training, or architecture. This leads to the central
empirical question of the paper: how much of current rejection is information loss, and how much lies
above it, and can this distinction guide an effective improvement?

\textbf{Our approach.} At each draft position, we compare a drafter with the best proposal subject to
the same information constraint. For example, suppose the target has two plausible continuations,
\texttt{of course} and \texttt{no problem}. An all-parallel drafter must fix its second-position
proposal before knowing whether the first sampled token is \texttt{of} or \texttt{no}, so it can
produce the mismatched block \texttt{of problem}. The rejection of the best proposal under this
constraint is the \textit{information floor}, which prices this unresolved branch choice. The
difference between the observed rejection and the floor is the
\textit{model gap}, the loss not forced by unavailable target realisations. This gives the decomposition
\[
\text{observed rejection}
=
\text{information floor}
+
\text{model gap}.
\]
\cref{sec:setup} gives the formal definitions.

Crucially, the information floor can be measured from target rollouts. At a given prompt, sampled
continuations reveal the alternative paths that a proposal under the chosen information constraint
must cover. Finding the best shared proposal across these paths estimates the floor, while evaluating
a drafter on the same prompts gives its model gap. Repeating this measurement across prompts, domains,
and target models lets us determine which source of rejection dominates in practice.

\textbf{Findings.} We measure information floors and model gaps across four domains using Qwen3-4B,
then repeat the analysis on Qwen3-8B and Qwen3-14B \citep{yang2025qwen3technicalreport}, and
Gemma-4-12B \citep{gemmateam2026gemma4technicalreport}. Because estimating a floor requires only
access to target probabilities, we also measure it on DeepSeek-V4-Pro
\citep{deepseekai2026deepseekv4highlyefficientmilliontoken} through its API. Three findings emerge.

\emph{Parallel blindness creates a substantial floor, especially for open-ended generation.} On
Qwen3-4B, the floor rises from zero at the first draft position to $0.286$ at the seventh, corresponding
to a maximum per-slot acceptance of $71\%$ under this information constraint. Across all four targets,
open-ended chat produces higher floors than constrained arithmetic or code, linking the cost of
parallel blindness to how many plausible directions a continuation can take
(\cref{sec:q1,sec:q4}).

\emph{One realised token removes almost all of this floor.} Allowing each draft position to observe
only the immediately preceding target token reduces the floor by $86$--$100\%$ across the measured
positions, leaving an information floor of at most $0.041$. Further conditioning can therefore
reduce the floor by at most this small residual. An independent mutual-information analysis
also finds that one predecessor retains most of the predictive information in the within-block
history (\cref{sec:q2}).

\emph{Released block drafters retain large gaps above their information floors.} For DFlash on
Qwen3-4B, the model gap accounts for $55$--$67\%$ of per-slot rejection at every position
affected by parallel blindness. For DSpark, an oracle-conditioned control supplies the realised
predecessor, which matches its own draft on paths that survive to the slot. Under this control, the gap
accounts for $89$--$100\%$, while its information floor is an order of magnitude smaller than the
all-parallel floor. Across all four targets, the DFlash model gap accounts for $43$--$64\%$ of
final-position rejection (\cref{sec:q3,sec:q4}).

\textbf{From measurement to an intervention.} The small order-1 floor and large remaining model
gap motivate improving how a drafter uses its predecessor, rather than increasing the conditioning
order. We introduce a prefix-attention head in which the predecessor and parallel drafter state
jointly query the committed prefix. This removes the vanilla head's restriction to a
context-independent predecessor correction. Combined with short convolution and explicit slot
embeddings, the resulting Qwen3-4B drafter raises mean serving accepted length from $4.75$ to
$4.94$, a $3.93\%$ gain over the released DSpark checkpoint across nine tasks. We report the
complete-system comparison and short-budget component results in \cref{sec:solution}.

\textbf{How to read these results.} The model gap measures rejection that is not forced by missing
earlier tokens, but a particular drafter may not be able to eliminate all of that loss
(\cref{app:arch}). Our measurements also average over all target rollouts. During serving,
later positions are reached only on paths whose earlier draft tokens were accepted, which favours
paths the drafter already handles well. We measure this serving effect separately in
\cref{sec:q5}.

\section{Measuring Information Floors and Model Gaps}\label{sec:setup}

We separate per-slot rejection into the minimum allowed by an information constraint and the model
gap above it. \cref{tab:notation} collects the notation.

\textbf{Target, block, and trajectory law.} Let $X$ be a context and
$Z=(Z_0,\ldots,Z_{\gamma-1})\in\mathcal V^\gamma$ a target-sampled block over a finite vocabulary
$\mathcal V$, indexed from zero. A drafter proposes $\gamma$ tokens, verified left to right.
For a token prefix $s$, let $p_T(\cdot\mid s)$ denote the target's next-token distribution.
The target block law at $X=x$ is
$\mu_x(z):=\prod_{i=0}^{\gamma-1}p_T(z_i\mid x,z_{<i})$, where $z_{<i}=(z_0,\ldots,z_{i-1})$.
Let $\nu$ be the stated context distribution. We use $\mu$ for the joint evaluation law
$X\sim\nu$, $Z\mid X\sim\mu_X$. Fixed-anchor expectations use $\mu_x=\mu(\cdot\mid X=x)$.

\textbf{Conditioning order and proposals.} At slot $k$, an order-$m$ proposal observes
$\Iset_{k,m}:=(X,Z_{k-m},\ldots,Z_{k-1})$, for $0\le m\le k$.
We write $\Iset_m$ when $k$ is fixed.
At $m=0$, only $X$ is available. Increasing $m$ reveals a longer suffix of the realised history.
For target $p_T$ and drafter rule $q_{D,k}$, define
\[
p_k(\cdot):=p_T(\cdot\mid X,Z_{<k})=p_T(\cdot\mid\Iset_{k,k}),\qquad
q_k(\cdot):=q_{D,k}(\cdot\mid\Iset_m).
\]
These are path-evaluated distributions under $\mu$. Subscripts $T$ and $D$ denote target and
drafter.

\textbf{Per-slot rejection risk.} For a target distribution $p$ and a proposal distribution $q$ on
$\mathcal V$, let $\TV(p,q):=\frac12\sum_{v\in\mathcal V}|p(v)-q(v)|$.
A token sampled from $q$ and verified against $p$ is accepted with probability
\begin{equation}
\alpha(p,q) \;=\; \sum_v \min\big(p(v),\,q(v)\big) \;=\; 1 - \TV(p,q).
\label{eq:alpha}
\end{equation}
The risk $R_k:=\E_\mu[\TV(p_k,q_k)]$ averages rejection over target rollouts, without
conditioning on earlier draft acceptance. We call this the free-rollout risk.

\textbf{Information floor.} We compare the drafter's risk with the best attainable under the same
information constraint. For conditioning order $m$, define
\begin{equation}
T_k^{(m)} \;:=\; \E_\mu\Big[\; \min_q\ \E_\mu\big[\TV(p_k,q) \,\big|\, \Iset_m\big]\Big].
\label{eq:floor}
\end{equation}
The minimum is over all probability distributions $q$ on $\mathcal V$, chosen separately for each
information state. The inner expectation averages over hidden paths compatible with $\Iset_m$.
Thus $T_k^{(m)}$ is the least per-slot risk
allowed by the information constraint, independent of a particular drafter.

\begin{figure}
\centering
\includegraphics[width=\linewidth]{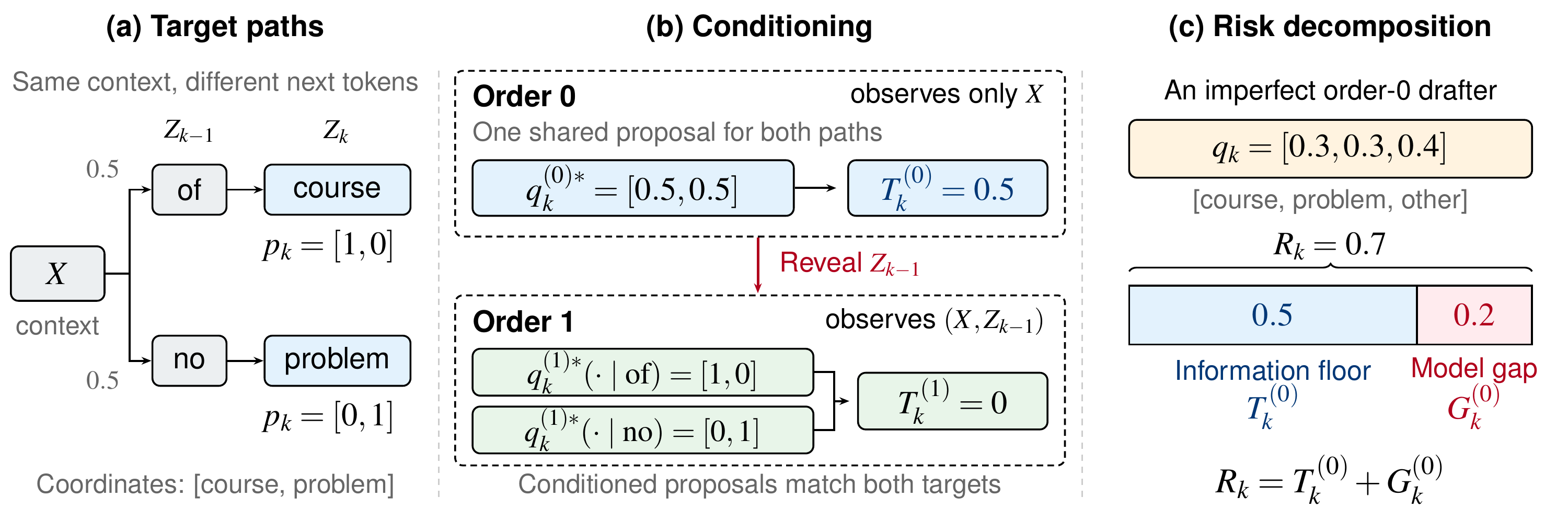}
\caption{Toy floor--gap decomposition at fixed $X$. Revealing $Z_{k-1}$ lowers the floor
from $0.5$ to $0$. Stars denote floor-attaining proposals, with fixed $X$ suppressed.}
\label{fig:overview}
\end{figure}

\textbf{Model gap and bounds.} By construction, every order-$m$ proposal satisfies
\begin{equation}
R_k \;\ge\; T_k^{(m)}, \qquad\text{and}\qquad T_k^{(0)} \;\ge\; \cdots \;\ge\; T_k^{(k)}=0 .
\label{eq:bound}
\end{equation}
More conditioning can only lower the floor, and full within-block history allows the proposal to
match the target exactly (\cref{app:derivations}). We define the order-$m$ model gap as
$G_k^{(m)} := R_k - T_k^{(m)}$. It measures rejection beyond what missing information forces.
Architectural constraints may make only part of this gap recoverable by a fixed drafter family
(\cref{app:arch}). \cref{fig:overview} illustrates the decomposition at a fixed context.

\textbf{Estimation.} An anchor is an evaluation position in a target-generated sequence, with its
preceding context $X$. At each anchor $X=x$, we draw $M$ target continuations to estimate the
expectations in \eqref{eq:floor} by sample averages. The empirical minimisation is a convex
optimisation over the $|\mathcal V|$ probabilities $q(v)$, solved by the algorithm in \cref{app:quantile}.
For slot $k$ and conditioning order $m$, we evaluate drafter risk on the same paths and form
$\widehat G_k^{(m)}(x)=\widehat R_k(x)-\widehat T_k^{(m)}(x)$, where hats denote estimates at the
fixed anchor. We aggregate anchors with Hájek weights and bootstrap prompts to preserve
dependence among their anchors. Full estimators and sensitivity analyses appear in
\cref{app:method,app:robust}.

\textbf{Accepted length and survival.} Serving additionally depends on whether earlier proposals are
accepted. Define the target-path weight
$a_i(X,Z_{\le i}):=\min(1,q_i(Z_i)/p_i(Z_i))$, abbreviated as $a_i$.
Since $Z_i$ is sampled from $p_i$, multiplying by $a_i$ recovers the accepted mass
$p_i(Z_i)a_i=\min(p_i(Z_i),q_i(Z_i))$ along the target path.
Let $W_{k-1}:=\prod_{i=0}^{k-1}a_i$, with $W_{-1}=1$. Its expectation is the probability of
reaching slot $k$. Let $J\in\{0,\ldots,\gamma\}$ count consecutive accepted drafts. Including a
correction or bonus token gives expected returned length
\begin{equation}
\tau := 1 + \E[J]
= 1 + \sum_{j=0}^{\gamma-1} \E_\mu\Big[\prod_{i=0}^{j} a_i\Big].
\label{eq:tau}
\end{equation}
Unlike $R_k$, $\tau$ retains the joint accept factors across positions. In \cref{sec:q5}, we use
$W_{k-1}$ to relate free-rollout risks to risks on reached paths.

\section{What the measurements reveal}\label{sec:findings}

Our diagnosis proceeds in three steps: measure the cost of parallel blindness, reveal one predecessor,
and compare existing drafters with the resulting floor.

\subsection{How much does parallel blindness cost?}\label{sec:q1}

\textbf{Default measurement setting.} Unless stated otherwise, the following measurements use
Qwen3-4B as the target, a block length of seven, and 384 anchors drawn from 170 prompts across gsm8k
\citep{cobbe2021trainingverifierssolvemath}, mbpp \citep{austin2021programsynthesislargelanguage},
alpaca \citep{alpaca}, and arena-hard \citep{li2024crowdsourceddatahighqualitybenchmarks}. Target
continuations are sampled at temperature 1 without top-$p$
truncation. Each anchor lies on a target-generated sequence, so the floors are measured at contexts
visited by the target itself. The floor curves in this section use
$M{=}256$ full-vocabulary rollouts per anchor.

\begin{figure}
\centering
\includegraphics[width=\linewidth]{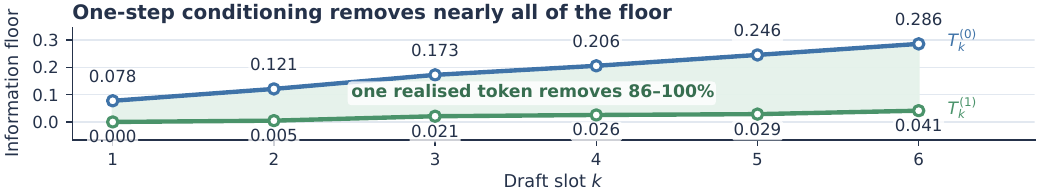}
\caption{Order-0 and order-1 information floors on the same $M{=}256$ free rollouts per anchor,
using the full vocabulary and exact total variation.}
\label{fig:t1}\label{fig:t0}
\end{figure}

\textbf{The floor is large and grows with depth.} In \cref{fig:t1}, the order-0 floor rises steadily from $0.078$ at
slot 1 to $0.286$ at slot 6. Equivalently, the information constraint caps the best attainable
per-slot acceptance of an ideal all-parallel drafter at about $92\%$ at slot 1 and $71\%$ at slot 6.
At slot 0, the floor is exactly zero because no within-block token has yet been realised. This
identity also serves as a check on our estimator.

\begin{figure}
\centering
\includegraphics[width=\linewidth]{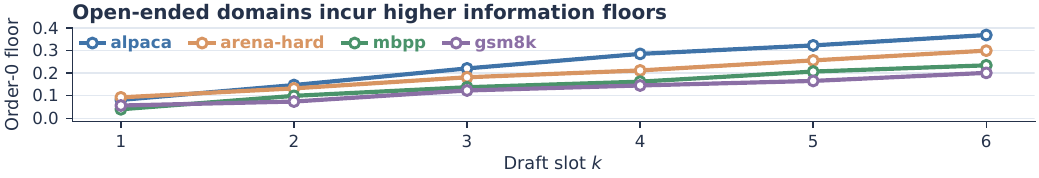}
\caption{Order-0 information floors by domain across the block.}
\label{fig:domain-floors}
\end{figure}

The cost of parallel blindness also depends on the domain. Splitting the order-0 floor in
\cref{fig:domain-floors} shows that, at slots 1--6, the two open-ended domains have higher floors than the two
constrained domains. Open-ended prompts support a wider range of plausible continuations, so the
target's realised path reveals more about the distribution at later slots. This pattern is consistent
with task constraints in gsm8k and mbpp narrowing this range and reducing the cost of parallel
blindness.

At the level of individual contexts, the cost is unevenly distributed. At slot 1, two thirds of anchors have a
floor below $0.01$ and together
contribute only $0.4\%$ of the total floor at that slot, whereas the top decile contributes $60\%$.
The distribution becomes less concentrated with depth but remains uneven: at slot 6, the mean of
$0.286$ exceeds the median of $0.199$, one quarter of anchors remain below $0.01$, and the top decile
contributes $27\%$.
The aggregate floor therefore combines contexts with little information loss and contexts where
missing within-block history imposes a substantial constraint.

\subsection{How much conditioning is needed?}\label{sec:q2}

Parallel blindness is costly, but the appropriate architectural response depends on how much realised
history is needed to remove that cost. We compare the order-0 floor with the order-1 floor, which lets
each slot condition only on the immediately preceding realised token. By \eqref{eq:bound}, the
residual captures everything that longer realised prefixes could still remove.

\textbf{One realised token removes almost all of the floor.} Conditioning on the preceding realised
token removes $86$--$100\%$ of the order-0 floor and leaves the order-1 floor at $0.041$ or below
at every measured slot (\cref{fig:t1}). In the acceptance metric, almost all of the
cost of parallel blindness is therefore removed by one step of conditioning.

We estimate both floors from the same target rollouts, grouping paths by their realised predecessor
for the order-1 estimate. Split-half fitting, numerical checks, and an independent implementation
support the comparison, as detailed in \cref{app:method,app:noise,app:twoimpl}.

\textbf{An independent information-theoretic test.} The TV floor measures locality through its effect
on speculative acceptance. We next ask how much of the predictive information in the full within-block
history is retained by a short suffix. Conditional mutual information provides a complementary view
under log loss. Under the target's own untruncated sampling law, we measure the fraction recovered by
the last $m$ realised tokens as
\[
\rho_{k,m}:=
\frac{I(Z_k;Z_{k-m},\ldots,Z_{k-1}\mid X)}
     {I(Z_k;Z_{<k}\mid X)},
\]
when the denominator is positive. Its log-loss derivation appears in \cref{app:ce}.

On anchors with a nontrivial information gap, one realised token recovers $94.3\%$ of the missing
path information, pooled across slots. Two tokens recover $99.0\%$, while extending the history to four yields little further
recovery. The same pattern holds across the measured positions, with slot-wise estimates and
intervals in \cref{tab:mi-locality} and estimator sensitivity in \cref{app:mi-sensitivity}.

The two analyses reach the same qualitative conclusion under different objectives. Mutual information
measures the predictive information retained from the realised history, whereas the TV floor measures
the minimum rejection risk under the corresponding information constraint. Despite using different
probes and anchor sets, both indicate that the immediately preceding token carries most of the useful
within-block information under the measured conditions.

Concentration of the target's continuations offers a possible explanation for this locality. At
slot 6, the six-token prefix distribution has a median effective support of $1.8$ trajectories,
indicating that probability is often concentrated on very few paths (\cref{app:concentration}).
When these likely paths have distinct predecessors, one token can identify which path was taken
and resolve uncertainty about its continuation. A small effective support alone does not establish
this condition, but it helps explain how a short history can suffice.

These results leave limited scope for reducing the information floor through longer histories. They
nevertheless leave open how accurately a drafter models the conditional distribution available from
one predecessor. We therefore next compare existing drafters with their corresponding information
floors.

\subsection{How far are deployed drafters from their floors?}\label{sec:q3}

We now compare real drafters with the floors imposed by their factorisations. We evaluate the paired
DFlash and DSpark checkpoints released with DSpark for Qwen3-4B. They share the backbone, data
pipeline, and optimiser schedule but differ in prediction head and training objective.

\textbf{The model gap dominates DFlash's rejection.} DFlash
\citep{chen2026dflashblockdiffusionflash} generates all block positions in one parallel pass without
within-block conditioning, so its information floor is $\Tz$. We estimate $R$ and $\Tz$ from the
same anchors, target paths, and conditionals, making
$G_k^{(0)}=R_k-T_k^{(0)}$ a paired estimate.

\begin{figure}
\centering
\includegraphics[width=\linewidth]{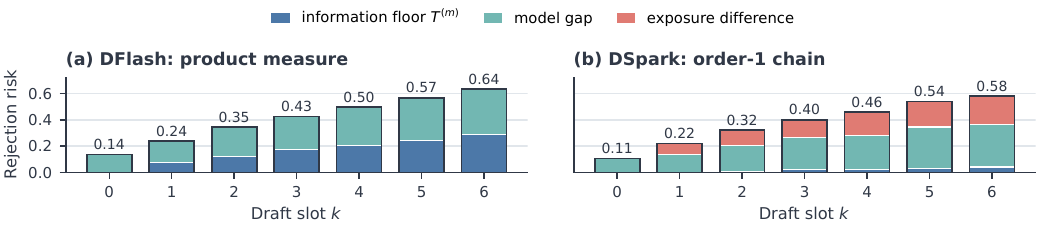}
\caption{Per-slot rejection-risk breakdowns on Qwen3-4B. For DFlash, $T^{(0)}$ and $G^{(0)}$ sum to
its risk $R$. For DSpark and $k\ge1$, $T^{(1)}+G^{(1)}=R^{\mathrm{oracle}}$ uses the target predecessor.
Adding $E^{\mathrm{expo}}$ gives the self-conditioned risk
$R^{\mathrm{self}}$. At slot 0, both use the order-0 decomposition.}
\label{fig:stack}
\end{figure}

In the left panel of \cref{fig:stack}, the gap accounts for $55$--$67\%$ of per-slot rejection at
slots 1--6. Slot 0 provides a boundary case: no within-block history is missing and its floor is zero,
yet the drafter still incurs rejection, all of which is model gap. Both components grow with depth,
but the gap remains larger than the floor throughout the block.

\textbf{DSpark remains far above its order-1 floor.} DSpark
\citep{cheng2026dsparkconfidencescheduledspeculativedecoding} adds a Markov head to the same parallel
backbone. Comparing it with $\To$ requires supplying the realised predecessor available to an
order-1 proposal. This control lets us assess how well the head models the conditional distribution
when that information is available. Using its own sampled predecessor would also introduce errors
from conditioning on a token that may differ from the target's realised path.

For $k\ge1$, feeding the head the target predecessor $Z_{k-1}$ defines the oracle-conditioned risk
$R_k^{\mathrm{oracle}}:=\E_\mu[\TV(p_k,q_{D,k}(\cdot\mid X,Z_{k-1}))]$
and the corresponding model gap $G_k^{(1)}:=R_k^{\mathrm{oracle}}-T_k^{(1)}$.
We estimate the floor and risk on the same anchors and target rollouts, with estimator details
and uncertainty intervals in \cref{app:drafter-details}.

In the right panel of \cref{fig:stack}, this gap accounts for $89$--$100\%$ of the oracle-conditioned
risk across slots 1--6. Thus, even with the correct predecessor, most of DSpark's rejection remains
above the information floor. The small order-1 floor does not translate into comparably low rejection
for the released drafter.

When the head instead uses its own sampled predecessors, rejection increases further. Recursively
sampling from slot 0 defines the self-conditioned risk $R_k^{\mathrm{self}}$. For $k\ge1$, we call
$E_k^{\mathrm{expo}}:=R_k^{\mathrm{self}}-R_k^{\mathrm{oracle}}$ the exposure difference, giving
\begin{equation}
R_k^{\mathrm{self}} \;=\; \underbrace{T_k^{(1)}}_{\text{floor}} \;+\;
\underbrace{G_k^{(1)}}_{\text{model gap at order 1}} \;+\;
\underbrace{E_k^{\mathrm{expo}}}_{\text{exposure difference}} .
\label{eq:threeway}
\end{equation}
At slot 0, no within-block predecessor exists, the applicable floor is
$T_0^{(0)}=0$, and the two risk evaluations coincide.
At slot 6, the exposure difference is $0.214$. This diagnoses sensitivity to predecessor errors.
It is not a serving loss, because standard left-to-right verification stops after an earlier rejection
and does not use subsequent drafts on that path. \cref{sec:q5} accounts for survival when relating
per-slot risks to serving.

\subsection{How do floors and gaps vary across targets and scale?}\label{sec:q4}

\begin{figure}
\centering
\includegraphics[width=0.8\linewidth]{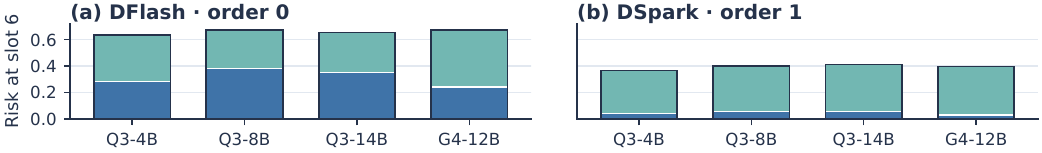}
\vspace{-0.5em}
\caption{Final-slot rejection-risk decompositions across four targets. Blue denotes the information
floor, teal the corresponding model gap, and the full bar height the corresponding measured risk.}
\label{fig:cross-target}
\end{figure}

\textbf{The gap persists across four targets.} To test whether the diagnosis extends beyond
Qwen3-4B, we repeat the pipeline on Qwen3-8B, Qwen3-14B, and Gemma-4-12B with their released
DFlash and DSpark drafters. All four targets use the same four domains and $384$ anchors each.
\cref{fig:cross-target} compares their final-slot floor--gap decompositions.

Both drafters remain well above their corresponding floors. For DFlash, the model gap accounts for
$43\%$--$64\%$ of final-slot risk. For DSpark, the order-1 floor remains small, leaving
$85\%$--$92\%$ of the oracle-conditioned risk in the model gap (\cref{app:method}). Even with the
realised predecessor, DSpark's risk exceeds the order-0 information floor $\Tz$ on every target.
The large gap between the information limit and the released drafter's risk therefore persists
across target models and both proposal structures.

\textbf{Frontier-scale evidence.} Extending the diagnosis to DeepSeek-V4-Pro requires a different
evaluation because its API exposes target probabilities but not the hidden states needed by a paired
block drafter. We estimate floors from its truncated probabilities, with omitted-mass bounds and
matched-path truncation checks in \cref{app:trunc}. Evidence about drafter rejection comes from
separate serving reports.

The measurements reproduce both the domain dependence and the locality observed on the open-weight
targets. Open-ended domains again have higher order-0 floors, and conditioning on one realised token
removes $86$--$90\%$ of the floor across slots. The remaining order-1 floor is about $0.03$ at the
final slot.

Published serving results for DeepSeek-V4-Pro-DSpark \citep{vllm2026dspark} separately give
approximate lower bounds of $0.11$ on batch-1 coding and $0.23$ on roleplay for the largest
per-slot rejection risk. These bounds follow from accepted length without assuming independent
acceptance across slots (\cref{app:frontier}). Their scale exceeds the small order-1 floor measured
through the API, consistent with the separation between information limits and drafter rejection
on the open-weight targets. Different workloads and sampling laws make this an indirect comparison
of scale, without a quantitative model-gap estimate.

\subsection{How does per-slot risk translate to serving performance?}\label{sec:q5}

Serving reaches slot $k$ only after earlier proposals are accepted. For $0\le k<\gamma$ with
$\Pr(J\ge k)>0$, the risk on reached paths is
\begin{equation}
R_k^{\mathrm{serve}}
:=
\Pr(J=k\mid J\ge k)
=\frac{\E_\mu\!\left[W_{k-1}\,\TV(p_k,q_k)\right]}
     {\E_\mu[W_{k-1}]}.
\label{eq:serving-risk}
\end{equation}
Here $J=k$ means rejection at slot $k$, and $W_{k-1}$ is the survival weight from \cref{sec:setup}.
For DSpark, the oracle-conditioned proposal applies because the realised predecessor equals its
accepted predecessor on reached paths.

\textbf{Survival selects paths the drafter handles well.} Reweighting lowers risk increasingly with
depth. At the final slot, risk falls from $0.635$ to $0.211$
for DFlash and from $0.366$ to $0.158$ for DSpark. Later slots retain paths favoured by earlier
acceptance decisions (\cref{tab:serving-risk-details}).
This selection reflects dependence among acceptances across slots. For both drafters, measured joint survival
exceeds the product of marginal acceptances. Treating slots as independent therefore underestimates
accepted length, as shown by the paired calculations in \cref{app:serving-details}.

\textbf{Larger model gaps need not yield larger serving gains.} To compare positions, an oracle fits
one distribution per prefix for DFlash, or per prefix and realised predecessor for DSpark.
For released proposal $q^{\mathrm{base}}$, we optimise $q_k$ with the same information while fixing
all other mappings $q_{-k}^{\mathrm{base}}$:
\begin{equation}
\Delta\tau_k^{\mathrm{BR}}
:=
\max_{q_k}\tau\!\left(q_k,q_{-k}^{\mathrm{base}}\right)
-\tau\!\left(q^{\mathrm{base}}\right),
\label{eq:slot-br}
\end{equation}
Because $q_k$ is optimised separately for every prefix, $\Delta\tau_k^{\mathrm{BR}}$ upper-bounds the
gain of any deployable slot-$k$ change with the same information and other slots fixed.

\begin{figure}
\centering
\includegraphics[width=\linewidth]{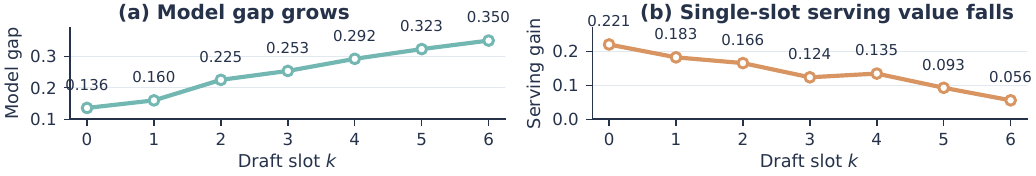}
\caption{DFlash's model gap rises with draft depth as single-slot serving value falls.}
\label{fig:slot-priority}
\end{figure}

For DFlash, $G_k^{(0)}$ grows by $2.6\times$ from slot 0 to 6, while the estimated gain
$\Delta\tau_k^{\mathrm{BR}}$ falls by $3.9\times$
(\cref{fig:slot-priority,app:single-slot-results,app:iterbr}). Slot 0 is always reached and can
preserve the entire suffix.
This is consistent with DSpark's interpretation that deeper parallel drafters can offset suffix decay
through stronger early predictions \citep{cheng2026dsparkconfidencescheduledspeculativedecoding}.

\section{From diagnosis to a prefix-attention head}\label{sec:solution}

One realised predecessor removes most of the information floor, yet existing order-1 drafters
remain far above it (\cref{sec:q2,sec:q3}). This motivates a
\emph{prefix-attention head}: the predecessor guides a fresh read of the committed prefix.
The design changes how the available information is used, without revealing more within-block history.
Following \cref{sec:q5}, we assess conditional modelling and serving gains separately.

\textbf{From a predecessor bias to a conditional prefix read.} Let $u_k(X)$ be the parallel
backbone's base logits and $a$ the preceding token. The vanilla DSpark head adds a low-rank
correction, \mbox{$\ell_k(X,a)=u_k(X)+B\,A[a]$}. Writing $B_v$ for row $v$ of $B$,
changing the predecessor from $a'$ to $a$ changes the log odds of successor $v$ versus $w$ by
\begin{equation}
(B_v-B_w)\big(A[a]-A[a']\big).
\label{eq:separable-head}
\end{equation}
The base logits cancel. Although the normalised proposal depends on both $X$ and $a$, the
predecessor-induced change in relative preference is independent of $X$. Increasing the quality
of the parallel backbone does not remove this additive restriction. The order-1 information
constraint itself permits richer interactions: the same predecessor can call for different
successors depending on the committed context.

Our head lets that change depend on the prefix. With drafter state $h_k(X)$, frozen target embedding $E(a)$,
and committed-prefix features $H_X$, it computes
\begin{equation}
x_{k,a}=f_{\mathrm{in}}\!\left([E(a);h_k(X)]\right),\qquad
\ell_k(X,a)=u_k(X)+B\,f_{\mathrm{head}}(x_{k,a},H_X).
\label{eq:prefix-head}
\end{equation}
Here $f_{\mathrm{in}}$ separately normalises the inputs and projects their concatenation.
$f_{\mathrm{head}}$ uses the predecessor-dependent query in prefix cross-attention followed by
a feed-forward transformation.

The head observes only the committed prefix and one predecessor. Its information floor remains
$T^{(1)}$ under the same measurement law, and zero at slot 0, where the predecessor is already
committed.

\textbf{Implementation and training.} To accommodate the heavier head, we adopt DFlash2's \citep{inco2026dflash2}
top-16 candidate scoring and lattice walk: the head scores candidate transitions across positions
in parallel, then a short walk selects the draft chain.

The backbone also uses a DFlash2-inspired two-tap convolution \citep{inco2026dflash2}, with
independent pre- and post-sublayer modules, plus learned slot embeddings. The differing convolution
wiring is detailed in \cref{app:solution-details}.

We retain DSpark's objectives and add a candidate-nomination loss to train the base logits to cover
likely target tokens. All terms use slot weights $\exp(-k/4)$, with loss definitions and weights
in \cref{app:solution-training}. We evaluate the resulting five-layer Qwen3-4B drafter after
10 epochs against the released DSpark checkpoint.

\begin{figure}[t]
\centering
\includegraphics[width=\linewidth]{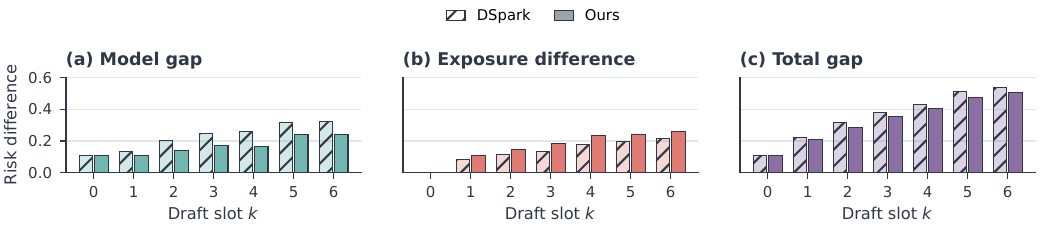}
\caption{Model gap, exposure difference, and their sum $R_k^{\mathrm{self}}-T_k^{(1)}$ for DSpark
(hatched) and ours (solid). Slot 0 uses the order-0 decomposition.}
\label{fig:solution-gap}
\end{figure}

\textbf{Model gap and exposure.} Under the same target-rollout protocol, the model gap
falls at slots 1--6, while slot 0 remains essentially unchanged
(\cref{fig:solution-gap,app:solution-gap}). At slot 6,
the gap falls by $25.5\%$. With the information floor unchanged, this reduction supports improved
use of the available predecessor and prefix.

The exposure difference increases, however, so the self-conditioned risk falls less than the
oracle-conditioned risk. Better predictions with the correct predecessor therefore do not imply
improved robustness to predecessor errors.

\textbf{Serving evaluation.} We evaluate both drafters in SGLang
\citep{zheng2024sglangefficientexecutionstructured} on 3,030 rows across nine tasks.
\Cref{tab:solution-serving} shows that acceptance gains translate into faster generation at both
temperatures despite the heavier head. This evaluates all architecture and training changes together
(\cref{app:solution-eval}).

One-epoch component comparisons (\cref{app:solution-variants}) support these design choices.
The head variant and short convolution each improve accepted length over the internal vanilla
baseline, slot embeddings alone have little effect, and the combined variant performs best.

\begin{table}
\centering
\caption{Task-macro accepted length ($\tau$) and speedup ($S$) over AR decoding on Qwen3-4B.}
\label{tab:solution-serving}
\small
\setlength{\tabcolsep}{14pt}
\begin{tabular}{lrrrr}
\toprule
 & \multicolumn{2}{c}{Temperature 0} & \multicolumn{2}{c}{Temperature 1} \\
\cmidrule(lr){2-3}\cmidrule(lr){4-5}
Drafter & $\tau$ & $S$ & $\tau$ & $S$ \\
\midrule
DSpark & 5.17 & $3.62\times$ & 4.75 & $3.34\times$ \\
Ours & \textbf{5.32} & $\mathbf{3.76}\times$ & \textbf{4.94} & $\mathbf{3.52}\times$ \\
\bottomrule
\end{tabular}
\end{table}

\section{Discussion and related work}\label{sec:disc}

\textbf{Speculative drafting spans several proposal structures.} Classical methods use a smaller
autoregressive drafter, with later work improving training, distillation, and adaptation
\citep{leviathan2023fast,chen2023acceleratinglargelanguagemodel,xia-etal-2023-speculative,
kim2023speculativedecodingbiglittle,ICLR2024_8766fbc6,liu2024onlinespeculativedecoding}.
Blockwise methods predict several tokens in parallel
\citep{NEURIPS2018_c4127b91,pmlr-v235-gloeckle24a,pmlr-v235-cai24b}. Sequential heads and
feature-autoregressive drafters restore dependence inside the proposal
\citep{ankner2024hydrasequentiallydependentdraftheads,pmlr-v235-li24bt,li2025eagle3scalinginferenceacceleration},
while multi-candidate methods organise alternative continuations for drafting and verification
\citep{NEURIPS2023_6034a661,Miao_2024,NEURIPS2024_ea1f5f08,li-etal-2024-eagle,inco2026dflash2,li2026chainstreesparentconditioneddrafting}.

\textbf{Conditioning history and proposal quality.} DSpark's similar results with
Markov and RNN heads \citep{cheng2026dsparkconfidencescheduledspeculativedecoding} are consistent
with the limited value of longer histories measured here. DeLS-Spec augments a frozen DFlash drafter
with an independently trained local head for within-block causal conditioning
\citep{zheng2026delsspecdecoupledlongshortcontexts}.
Domino and xPress explore causal refinement of proposals
\citep{huang2026dominodecouplingcausalmodeling,wang2026xpressparallelrefinementdiffusion}.
Our decomposition helps distinguish gains from additional information and gains from modelling
the available information more accurately.

\textbf{Serving value guides effort across positions.} DFlash improves with deeper drafters and
target features above an unchanged product floor \citep{chen2026dflashblockdiffusionflash}. Its decaying
position weights and D-PACE's dynamic weights \citep{wu2026dpace} likewise support prioritising early
slots by prefix reach and downstream value.

Taken together, the measurements favour improving how drafters use a short realised history: one
predecessor removes most of the information floor, but substantial model gaps remain. Our
prefix-attention intervention illustrates this direction. Serving gains additionally depend on
prefix survival and downstream acceptance, making position-specific value central to improvement.

\clearpage
\subsection*{AI use statement}

We used generative AI tools, including OpenAI Codex and Anthropic Claude Code, to help structure and
edit the manuscript; locate and summarise related literature; improve figures and tables; provide
feedback on the measurement methodology and robustness checks; assist with the presentation and
checking of mathematical derivations; implement, execute, and monitor parts of the experimental
pipeline; interpret empirical results; check consistency among notation, claims, and reported values;
and edit analysis, plotting, and document-preparation code. The authors defined the research questions
and experimental objectives, and retained final control over all research decisions. The
authors take responsibility for the final content of the paper.

\bibliography{ref}
\bibliographystyle{iclr2027_conference}
\clearpage
\appendix
\crefalias{section}{appendix}
\crefalias{subsection}{appendix}
\crefalias{subsubsection}{appendix}

\section{Properties of the information floor}\label{app:derivations}

\subsection{Notation}

\cref{tab:notation} summarises the notation used throughout the paper.

\begin{table}
\centering\small
\caption{Notation used throughout the paper.}
\label{tab:notation}
\begin{tabular}{ll}
\toprule
$X$, $Z$, $\gamma$ & context; target block; block length ($\gamma=7$ unless stated otherwise) \\
$\nu$, $\mu$, $\mu_x$ & context law; joint law of $(X,Z)$; target-path law given $X=x$ \\
$p_k$, $q_k$ & target and proposal distributions evaluated on a target path \\
$\alpha(p,q)$, $R_k$ & pairwise acceptance; free-rollout risk $\E_\mu[\TV(p_k,q_k)]$ \\
$a_i$, $W_{k-1}$ & target-path weight $\min(1,q_i(Z_i)/p_i(Z_i))$; reach $\prod_{i=0}^{k-1}a_i$ \\
$J$, $\tau$ & accepted draft-prefix length; expected returned length $1+\E[J]$ \\
$\Iset_m$ & shorthand for $\Iset_{k,m}=(X,Z_{k-m},\ldots,Z_{k-1})$ at fixed $k$ \\
$T_k^{(m)}$ & information floor: the least risk any order-$m$ proposal can achieve \\
$G_k^{(m)}$ & model gap $R_k - T_k^{(m)}$: risk not forced by the information restriction \\
$\widehat R_k$, $\widehat T_k^{(m)}$, $\widehat G_k^{(m)}$ & sample estimates; $(x)$ denotes evaluation at a fixed anchor $X=x$ \\
$R_k^{\mathrm{oracle}}$, $R_k^{\mathrm{self}}$ & target-predecessor risk; own-sampled-predecessor risk \\
$E_k^{\mathrm{expo}}$ & exposure difference $R_k^{\mathrm{self}} - R_k^{\mathrm{oracle}}$ \\
$M$ & rollouts per anchor, specified for each probe \\
\bottomrule
\end{tabular}
\end{table}

\subsection{The floor bounds acceptance}

For any proposal $q$ measurable with respect to $\Iset_m$, the definition of per-slot risk gives
$R_k = \E_\mu[\TV(p_k,q)]$. Conditioning on $\Iset_m$ and applying the tower rule yields
\[
R_k = \E_\mu\!\left[
  \E_\mu\!\left[\TV\!\left(p_k,q(\cdot\mid\Iset_m)\right)\mid\Iset_m\right]
\right].
\]
The inner expectation is at least its minimum over $q(\cdot\mid\Iset_m)$, which proves
$R_k\ge T_k^{(m)}$. Monotonicity follows because $\Iset_m$ refines $\Iset_{m-1}$: a proposal with
more information can always ignore it. At $m=k$, the proposal observes the entire realised
within-block prefix $Z_{<k}$, so each conditional family is a singleton and $T_k^{(k)}=0$.

\subsection{The log-loss reading: \texorpdfstring{$\E[\Delta\mathrm{CE}] = I(Z_k; Z_{<k}\mid X)$}{E[dCE] = I(Zk;Z<k|X)}}\label{app:ce}

We compare the realised token's log loss for an observer that knows the realised within-block
prefix and one that knows only the block input:
\[
C_{k,k}=-\log \Pr_\mu(Z_k\mid X,Z_{<k}),
\qquad
C_{k,0}=-\log \Pr_\mu(Z_k\mid X).
\]
The blind distribution is obtained by marginalising the unknown prefix. For any fixed vocabulary
item $y$, the law of total probability gives
\begin{equation}
\Pr_\mu(Z_k=y\mid X)
=\sum_{z\in\mathcal V^k}\Pr_\mu(Z_{<k}=z\mid X)\,p_T(y\mid X,z)
=\E_\mu\!\left[p_T(y\mid X,Z_{<k})\mid X\right].
\label{eq:blind-marginal}
\end{equation}
The expectation marginalises over a fresh prefix while holding $y$ fixed. After this distribution is
formed, it scores the realised random token $Z_k$.

The two entropy equalities now follow directly from the definition of conditional entropy,
\[
H(U\mid V):=\E_{U,V}\!\left[-\log p(U\mid V)\right].
\]
All variables are drawn from the target's joint sampling law, so
\[
\E[C_{k,0}]=H(Z_k\mid X),
\qquad
\E[C_{k,k}]=H(Z_k\mid X,Z_{<k}).
\]
Subtracting them yields
\begin{equation*}
\E[C_{k,0}-C_{k,k}]
=H(Z_k\mid X)-H(Z_k\mid X,Z_{<k})
=I(Z_k;Z_{<k}\mid X).
\end{equation*}
For example, suppose the realised prefix is equally likely to end in \texttt{of} or \texttt{no},
followed deterministically by \texttt{course} or \texttt{problem}. The full observer assigns the
realised next token probability one, while the blind observer assigns it probability one half. The
missing prefix therefore adds $\log 2$ nats to the token's negative log-probability, exactly the
information identifying which branch was realised.

In the estimator, $C_{k,k}$ is read directly from the teacher-forced probability along the
realised path, while \eqref{eq:blind-marginal} is estimated by averaging over sampled prefixes.
The identity survives truncation provided the trajectory law and the scoring conditional are the
\emph{same} truncated autoregressive law; what truncation destroys is the reading of the result as the
mutual information of the raw, unwarped target, which is the reading we want.

\section{Experimental setup and estimators}\label{app:method}

\subsection{Targets, drafters, corpora}

The primary target is Qwen3-4B in bf16 for \crefrange{sec:q1}{sec:q3} and
\cref{sec:q5}. \cref{sec:q4} repeats the floor-and-gap pipeline on Qwen3-8B,
Qwen3-14B, and Gemma-4-12B with the drafter released for each target, and measures
target-only frontier controls through the DeepSeek-V4-Pro API. On Qwen3-4B, we evaluate the official
\texttt{deepseek-ai/dflash\_qwen3\_4b\_block7} and
\texttt{deepseek-ai/dspark\_qwen3\_4b\_block7} checkpoints. Our target-side measurements use block
size $\gamma=7$, matching these checkpoints.

The four domains are gsm8k (grade-school arithmetic), mbpp (short Python),
alpaca (open instruction following), and an arena-hard long-context chat subset.
For arena-hard, we take up to the first 1024 distinct prompts in the released Arena-Hard-v2 order;
the frozen prompt identifiers are included with the measurement code.
The last has median context length $2250$ tokens, p90 $8226$, and maximum $8638$, compared with
medians near $440$ for the other three. The higher floors on the two open-ended domains are
consistent with task constraints narrowing the target's continuation family (\cref{sec:q1}).

\subsection{Anchors, aggregation and uncertainty}

An \emph{anchor} is a (prompt, position) pair at which a block of $\gamma$ tokens is measured.
We divide candidate anchors into 18 groups: six context-length buckets crossed with three response
positions (early, middle, and late). The context buckets are $[0,512)$, $[512,2048)$,
$[2048,4096)$, $[4096,8192)$, $[8192,16384)$, and $[16384,\infty)$ tokens; response positions are
the first, middle, and final thirds of the generated response. We sample within each group and record the inclusion probability
$\pi_i$ of every selected anchor. Except for the mutual-information recovery diagnostic described
below, for a selected-anchor set $S$ and anchor statistic $y_i$, we aggregate with the Hájek ratio
\[
\frac{\sum_{i\in S}y_i/\pi_i}{\sum_{i\in S}1/\pi_i},
\]
which normalises the inverse-inclusion weights over the realised sample. The design allows multiple
and overlapping anchors from one sequence, preserving the intended block-weighted estimand.

We obtain uncertainty intervals by resampling the observed prompts with replacement. Whenever a
prompt is selected, all its anchors are selected together, and we recompute the statistic on that
resample. Repeating this procedure $B=10^4$ times gives an empirical distribution; its 2.5th and
97.5th percentiles form the reported $95\%$ interval. Keeping each prompt's anchors together
preserves their dependence. This procedure reuses the recorded measurements and requires no new
target rollouts.

The headline configuration contains 96 anchors per domain, or 384 anchors over 170 prompts. The
floor and drafter probes use the same anchors, so $\widehat G_k^{(m)}$ is computed within each anchor.
Estimator-specific coverage and effective-sample-size gates are reported separately. All sampling
uses base seed 20260818, with deterministic per-prompt seeds derived from the prompt identifier.

Measuring $\rho_{k,m}$ requires many conditioned-prefix evaluations, so we do not run it on every
candidate anchor. An independent screening pass first identifies anchors whose final-slot missing
information exceeds $0.05$ nats. Within each domain, all eligible anchors fit within the measurement
budget, so all 698 receive the full recovery measurement, which re-estimates the denominator from
fresh paths. Their aggregation weights therefore use the original anchor inclusion probabilities $\pi_i$.
The prefix draws used to form the blind and partially conditioned distributions are independent of
the realised target path that supplies the scored token.

For anchor $i$, let $C_{k,m,i}$ be the estimated mean of $C_{k,m}$ at slot $k$, and define
$n_{i,k,m}:=C_{k,0,i}-C_{k,m,i}$ and $d_{i,k}:=C_{k,0,i}-C_{k,k,i}$ as the recovered and total
missing information. At each slot, we estimate
\[
\widehat\rho_{k,m}
=\frac{\sum_i n_{i,k,m}/\pi_i}{\sum_i d_{i,k}/\pi_i},
\]
where the sums run over retained eligible anchors. Pooling the same numerator and denominator over
eligible anchor--slot cells gives the pooled estimate. Thus $\widehat\rho_{k,m}$ measures the
fraction of missing information recovered over the sampled population. Anchors with little missing information make a correspondingly small contribution,
avoiding unstable averages of the per-anchor ratios $n_{i,k,m}/d_{i,k}$.

\subsection{Target rollouts and conditional families}

Unless stated otherwise, we sample from the target at temperature 1 without top-$p$ or top-$k$
truncation, with thinking disabled. Prompts use each target tokenizer's chat template with a generation
prompt appended. For each anchor, we draw $M$ continuations from a shared prefix.
At slot $k$, rollout $r$ provides its realised prefix $Z_{<k}^{(r)}$ and target distribution
\[
p_k^{(r)}=p_T\!\left(\cdot\mid X,Z_{<k}^{(r)}\right).
\]

For $T_k^{(0)}$, the empirical family contains all $M$ distributions with weight $1/M$. For
$T_k^{(1)}$, the grouping route partitions the same paths by their realised predecessor
$Z_{k-1}$. Within each group, we fit the floor on one half of the paths and score it on the other,
then average the group floors using their empirical frequencies. This split-half estimate is defined
on the path mass belonging to groups large enough to divide. The drafter decomposition reports this
covered mass explicitly.

Reusing the same paths to fit and score $T_k^{(0)}$ tends to underestimate that floor, while
split-half evaluation tends to overestimate $T_k^{(1)}$. Both effects reduce
$1-T_k^{(1)}/T_k^{(0)}$, so the reported locality is conservative.

A second route estimates the same floor without grouping rollouts that happen to share a suffix.
Fix the last $m$ realised tokens $z^\star=Z_{k-m:k-1}$ and denote the earlier path by
$S=Z_{<k-m}$. Conditional on $z^\star$, the floor is
\[
T_k^{(m)}(X,z^\star)
=
\min_q
\E_{S\sim \mu(\cdot\mid X,z^\star)}
\!\left[\TV\!\left(p_T(\cdot\mid X,S,z^\star),q\right)\right].
\]
Ordinary target rollouts sample $S$ from its marginal law given $X$, not from the conditional distribution above.
Bayes' rule relates the two:
\[
\Pr_\mu(S\mid X,z^\star)
\propto
\Pr_\mu(S\mid X)\,\Pr_\mu(z^\star\mid X,S).
\]
The sampled frequency of each path already represents the first factor. We therefore weight each
draw $S_r$ from the marginal law given $X$ by the likelihood of the observed suffix,
\[
\omega_r=\Pr_\mu(z^\star\mid X,S_r),
\qquad
\widetilde w_r=\frac{\omega_r}{\sum_s\omega_s}.
\]
After appending $z^\star$ to each sampled path, we evaluate
$p_k^{(r)}=p_T(\cdot\mid X,S_r,z^\star)$. The resulting estimate is
\[
\widehat T_k^{(m)}(X,z^\star)
=
\min_q\sum_r\widetilde w_r\,\TV(p_k^{(r)},q),
\]
which is the weighted TV problem solved below. We discard cells with insufficient effective sample
size; the path-count and gating checks appear in \cref{app:robust}.

\subsection{The TV barycentre solver}\label{app:quantile}

The rollout procedures above produce target distributions $p_k^{(1)},\ldots,p_k^{(M)}$ with
normalised weights $\widetilde w_1,\ldots,\widetilde w_M$. Minimising over probability distributions $q$ on the target
vocabulary gives the empirical floor
\begin{equation}
\widehat T
=
\min_q
\frac12\sum_v\sum_{r=1}^M
\widetilde w_r\,|p_k^{(r)}(v)-q(v)|.
\label{eq:weighted-barycentre}
\end{equation}
Order 0 uses $\widetilde w_r=1/M$. Grouping applies the same objective within each predecessor
group. SNIS supplies the nonuniform weights defined above.

Let $\widehat q^\star$ be a minimiser in \eqref{eq:weighted-barycentre}. The objective separates
across vocabulary coordinates except for the constraint
$\sum_v q(v)=1$. Define
\[
f_v(x)=\frac12\sum_r\widetilde w_r\,|p_k^{(r)}(v)-x|,
\]
and let $W_v^{<}(x)$ and $W_v^{=}(x)$ be the total weights of samples below and equal to $x$,
respectively. Its subgradient is
\[
\partial f_v(x)
=
\left[
W_v^{<}(x)-\frac12,\,
W_v^{<}(x)+W_v^{=}(x)-\frac12
\right].
\]
Let $\eta$ be the multiplier for the unit-mass constraint and define
$\beta=\frac12-\eta$. At every positive coordinate, the KKT condition becomes
\[
W_v^{<}\!\left(\widehat q^\star(v)\right)
\le \beta \le
W_v^{<}\!\left(\widehat q^\star(v)\right)+W_v^{=}\!\left(\widehat q^\star(v)\right).
\]
Thus every coordinate of $\widehat q^\star$ is a weighted $\beta$-quantile of
$\{p_k^{(r)}(v)\}_{r=1}^M$, and the same level $\beta$ is shared across the vocabulary. The unit-mass
constraint selects a feasible vector at that level.

The solver implements this characterisation directly. For each vocabulary coordinate $v$, let
$\sigma_v$ order the path values,
\[
p_k^{(\sigma_v(1))}(v)\le\cdots\le p_k^{(\sigma_v(M))}(v),
\qquad
C_{v,j}=\sum_{\ell=1}^j\widetilde w_{\sigma_v(\ell)}.
\]
The weights move with their paths; they do not change the ordering. The weighted $\beta$-quantile is
the first sorted value for which $C_{v,j}\ge\beta$. The solver reads this quantile in every
coordinate and sums the resulting coordinates. This sum is non-decreasing in $\beta$, so it searches
for the common level at which the sum crosses one. If the crossing falls inside a quantile interval,
it selects a point in that interval that makes the coordinates sum exactly to one. Thus
\begin{equation}
\widehat q^\star(v)
=
\operatorname{Quantile}_{\beta}
\!\left(\{p_k^{(r)}(v)\}_r;\{\widetilde w_r\}_r\right),
\qquad
\widehat T
=
\sum_r\widetilde w_r\,\TV\!\left(p_k^{(r)},\widehat q^\star\right).
\label{eq:wquantile}
\end{equation}
Uniform weights, used for order 0 and within each predecessor group, reduce cumulative weight to
ordinary sample counts. SNIS uses the same solver with the nonuniform weights defined above. Ties
can make the minimiser non-unique, but all valid minimisers attain the same floor value.

\subsection{Measuring drafter risk}

For each rollout $r$, we compare the target distribution $p_k^{(r)}$ with the drafter proposal
$q_k^{(r)}$ available under the same information. DFlash emits
$q_k^{(r)}=q_{D,k}(\cdot\mid X)$ per anchor and slot. For the DSpark control and $k\ge1$, we evaluate
$q_k^{(r)}=q_{D,k}(\cdot\mid X,Z_{k-1}^{(r)})$ using each target path's realised predecessor; at
slot 0, its proposal is $q_{D,0}(\cdot\mid X)$. In either case, the empirical average is
\[
\frac1M\sum_{r=1}^M
\TV\!\left(p_k^{(r)},q_k^{(r)}\right).
\]
For DFlash, this average is $\widehat R_k$. For the DSpark control it is
$\widehat R_k^{\mathrm{oracle}}$ when $k\ge1$; at slot 0, the oracle and self evaluations both equal
$\widehat R_0$. To obtain $\widehat R_k^{\mathrm{self}}$, replace the target predecessor in the
DSpark proposal by one obtained from a recursive drafter rollout given $X$.
We pair the DFlash and oracle risks with the order-0 and order-1 floor estimates, respectively, to
form the reported gaps; the self--oracle difference gives the exposure difference.

\subsection{Effective support}\label{app:effective-support}

For the concentration analysis in \cref{sec:q2,app:concentration}, let $n_c$ be the number of rollouts with realised
prefix $c=Z_{<k}$. We estimate the collision probability by
\[
\widehat\lambda
=
\frac{\sum_c n_c(n_c-1)}{M(M-1)},
\]
where $n_c(n_c-1)$ counts ordered pairs of distinct rollouts with prefix $c$. Equivalently,
\[
\widehat\lambda
=
\frac{1}{M(M-1)}\sum_{r\ne s}\mathbf 1\{Z_{<k}^{(r)}=Z_{<k}^{(s)}\}.
\]
For two independent rollouts from $\mu_x$,
\[
\E[\widehat\lambda]
=
\sum_c \Pr_{\mu_x}(Z_{<k}=c)^2,
\]
so the collision-probability estimate is unbiased. We report the collision-equivalent effective support
$N_{\mathrm{eff}}^{(2)}=1/\widehat\lambda$ when at least one pair collides. When no pair collides, we
record $M$ as a finite convention. We use this quantity as a descriptive
concentration diagnostic.

\section{Estimator validation and sensitivity}\label{app:robust}

We examine numerical resolution, agreement between independent implementations, finite-sample
sensitivity, vocabulary truncation, and sampling-law sensitivity. We then report uncertainty
intervals for the headline quantities.

\subsection{Numerical resolution}\label{app:noise}

The order-1 floor at slot 1 is zero by identity: conditioning on $Z_0$ fixes the entire realised
prefix relevant to that slot. The full-vocabulary grouping implementation returns $\ensuremath{1\times10^{-3}}$, a small
residual from batched bf16 arithmetic. We therefore treat approximately $10^{-3}$ as the numerical
resolution of this estimator. The deep-slot order-1 floors are $20$--$40$ times larger, while the
slot-2 floor is about five times larger and should be read at that precision.

\subsection{Independent replication}\label{app:twoimpl}

The headline estimates use a local full-vocabulary run with $M=256$ paths and predecessor
partitioning for the order-1 floor. We repeat the measurement through a separately implemented
\texttt{sglang} pipeline, using top-256 probabilities, $M=1024$ paths, and importance reweighting for
the order-1 floor.

On the 384 shared anchors, the order-0 estimates differ by $\ensuremath{7\times10^{-4}}$--$\ensuremath{3.4\times10^{-3}}$ across slots. At slot 6,
the paired difference between the order-1 estimates is $+\ensuremath{3.1\times10^{-3}}$, with a $95\%$ prompt-bootstrap
interval of $[-8.56,+13.90]\times10^{-3}$; the difference is at most $0.007$ throughout the block. The main
decompositions use the full-vocabulary run, in which the order-0 floor, order-1 floor, and drafter
risk are evaluated on the same target paths.

\begin{center}\small
\setlength{\tabcolsep}{4pt}
\begin{tabular}{lccccccc}
\toprule
slot $k$ & 0 & 1 & 2 & 3 & 4 & 5 & 6 \\
\midrule
$T_k^{(0)}$, full vocabulary, $M{=}256$ & 0.00 & 0.08 & 0.12 & 0.17 & 0.21 & 0.25 & 0.29 \\
$T_k^{(0)}$, top-256, $M{=}1024$ & 0.00 & 0.08 & 0.12 & 0.17 & 0.20 & 0.25 & 0.29 \\
\bottomrule
\end{tabular}
\end{center}

\subsection{Finite-sample sensitivity}

We compare $M=256$ with $M=1024$ on identical anchors under the same engine and sampling law.
The table reports absolute differences in units of $10^{-3}$:

\begin{center}\small
\begin{tabular}{lcccccc}
\toprule
slot & 1 & 2 & 3 & 4 & 5 & 6 \\
\midrule
$T^{(0)}$: $|M{=}1024-M{=}256|$ & 1.32 & 0.83 & 2.40 & 2.14 & 1.21 & --- \\
$T^{(1)}$: $|M{=}1024-M{=}256|$ & 0.00 & 0.14 & 1.21 & 1.86 & 0.11 & 0.08 \\
\bottomrule
\end{tabular}
\end{center}

Both floors move by at most $\ensuremath{2.4\times10^{-3}}$ under the fourfold change in path count. The order-0
$M=256$ run did not record slot 6, so that cell has no path-count-only comparison. A separate
full-vocabulary $M=256$ estimate differs from the top-256 $M=1024$ estimate by $\ensuremath{7\times10^{-4}}$, although
that comparison also changes implementation. At $M=1024$, the importance-sampling effective sample
size ranges from 688 to 924 across order-1 slots, and the slot-6 fit-score difference decreases from
$+6.6\times10^{-3}$ at $M=256$ to $+\ensuremath{2\times10^{-3}}$.

Split-half scoring provides a separate fit check for the order-0 solver. Fitting $q^\star$ on half
the paths and scoring it on the other changes $T^{(0)}$ by at most $0.006$ at any slot and by
$0.003$ after pooling, with no consistent sign. Its effective sample size is exactly $M$ because
order 0 uses uniform path weights.

\subsection{Longer blocks}\label{app:blocklen}

The reported measurements use a block length of seven. To check that the locality of \cref{sec:q2}
is not an artefact of that choice, we repeat the order-0 and order-1 measurement at $\gamma=16$ on
the replication pipeline of \cref{app:twoimpl} ($M=1024$, top-256), with the same per-anchor seeds,
so slots 0--6 repeat that run's estimates.
The anchors are the 372 of the original 384 that leave room for a sixteen-token block, with the same
strata and the same median context length.

\begin{center}\small
\setlength{\tabcolsep}{3.5pt}
\begin{tabular}{lcccccccc}
\toprule
slot $k$ & 0 & 1 & 2 & 3 & 4 & 5 & 6 & 7 \\
\midrule
$T_k^{(0)}$ & 0.00 & 0.08 & 0.12 & 0.17 & 0.20 & 0.24 & 0.29 & 0.31 \\
$T_k^{(1)}$ & --- & 0.00 & \ensuremath{2.5\times10^{-3}} & 0.02 & 0.02 & 0.03 & 0.04 & 0.04 \\
\midrule
slot $k$ & 8 & 9 & 10 & 11 & 12 & 13 & 14 & 15 \\
\midrule
$T_k^{(0)}$ & 0.33 & 0.37 & 0.39 & 0.41 & 0.43 & 0.44 & 0.46 & 0.47 \\
$T_k^{(1)}$ & 0.04 & 0.05 & 0.04 & 0.04 & 0.06 & 0.06 & 0.05 & 0.09 \\
\bottomrule
\end{tabular}
\end{center}

The order-0 floor keeps growing with depth, reaching $0.47$ at slot 15 and capping the per-slot
acceptance of an ideal all-parallel drafter at $52.7\%$. The order-1 floor also grows but stays at
$0.090$ or below, so one realised token removes $80.9$--$97.9\%$ of the floor across slots 2--15.
The removed share is lowest at the final slot, where the order-1 floor is largest. Extending the
block from seven to sixteen slots therefore raises the final-slot floor from $0.286$ to $0.473$,
while a single realised token still removes most of it. The per-domain ordering of
\cref{sec:q1} also persists: at slot 15 the floors are $0.619$ on alpaca and $0.479$ on arena-hard
against $0.389$ on mbpp and $0.377$ on gsm8k.

\subsection{Mutual-information recovery sensitivity}\label{app:mi-sensitivity}

The headline recovery fractions use $M=512$ paths and retain all eligible anchor--slot cells.
Applying effective-sample-size thresholds of 32 and $M/2$ to the same measurements gives pooled
order-1 estimates ranging from $93.9\%$ to $95.2\%$ across the three thresholds. The pooled order-2
and order-4 estimates vary by less than $0.4$ percentage points.
Under the matched relative threshold, increasing $M$ from 64 to 512 moves the pooled order-1 estimate from
$92.7\%$ to $93.9\%$.

\subsection{Vocabulary truncation}\label{app:trunc}

For a top-$K$ probability read, let $r_Z$ be the target mass outside the returned support. After
renormalising the retained probabilities, write $\widetilde T$ for the resulting floor. Then
$|T-\widetilde T|\le\E_\mu[r_Z]$. We report this residual band with every truncated estimate.

The frontier-scale floor in \cref{sec:q4} is read through an API capped at the top 20 tokens. We
calibrate this read on the same 384 Qwen3-4B anchors and sampled paths, using $M=256$ for both
top-20 and top-256:

\begin{center}\scriptsize
\setlength{\tabcolsep}{2pt}
\begin{tabular}{lcccccc}
\toprule
slot & 1 & 2 & 3 & 4 & 5 & 6 \\
\midrule
paired $T_k^{(0)}$ difference (top-20 $-$ top-256) & $-1.4{\cdot}10^{-7}$ & $+1.4{\cdot}10^{-5}$ & $+2.5{\cdot}10^{-6}$ & $+1.6{\cdot}10^{-5}$ & $+1.6{\cdot}10^{-5}$ & $+1.0{\cdot}10^{-5}$ \\
\bottomrule
\end{tabular}
\end{center}

The H\'ajek-weighted absolute difference is at most $1.7\times10^{-5}$ across the block. This
calibration supports the top-20 frontier measurement used in \cref{sec:q4}.

The frontier API sample contains 120 anchors. Of these, 118 have a complete slot-6 record and
contribute to the reported $T_6^{(0)}=0.25$. The held-out order-1 estimate at that slot is about
$0.03$, giving the absolute scale used in \cref{sec:q4}. The mean omitted mass is $2.2\times10^{-4}$.

\subsection{Sampling-law sensitivity}

We repeat the gsm8k comparison under the drafter training law: temperature $0.7$, top-$p$ $0.8$, and
top-$k$ $20$. The trajectory and verification distributions use the same transformed law. The
baseline uses temperature one without truncation; anchors and prefixes are fixed across the two
conditions.

\begin{center}\small
\begin{tabular}{lcccccccc}
\toprule
slot & $T$ base & $T$ train & $R$ base & $R$ train & $G^{(0)}$ base & $G^{(0)}$ train & \multicolumn{2}{c}{$G^{(0)}/R$} \\
\midrule
0 & 0 & 0 & 0.02 & 0.02 & 0.02 & 0.02 & 100\% & $\to$ 100\% \\
3 & 0.09 & 0.06 & 0.19 & 0.18 & 0.10 & 0.12 & 52\% & $\to$ 69\% \\
5 & 0.15 & 0.09 & 0.26 & 0.22 & 0.11 & 0.13 & 43\% & $\to$ 58\% \\
6 & 0.16 & 0.09 & 0.32 & 0.29 & 0.16 & 0.19 & 49\% & $\to$ 67\% \\
\bottomrule
\end{tabular}
\end{center}

At slot 6, truncation reduces the information floor by $42\%$ and drafter risk by $11\%$, increasing
$G^{(0)}/R$ to $67\%$. The model gap therefore remains dominant under this sampling law. At the baseline
law, the two proposal-warping implementations coincide; under the training law, their risks differ by
at most $0.008$.

\subsection{Headline uncertainty intervals}

\begin{table}
\centering\scriptsize
\caption{95\% prompt-bootstrap confidence intervals for the headline floor estimates. Daggers mark
exact slot-1 identities; the raw numerical residual is reported below.}
\label{tab:floor-intervals}
\setlength{\tabcolsep}{4pt}
\begin{tabular}{lcccccc}
\toprule
slot $k$ & 1 & 2 & 3 & 4 & 5 & 6 \\
\midrule
$T_k^{(0)}$ (top-256) & [.055,.101] & [.092,.149] & [.136,.211] & [.164,.246] & [.202,.292] & [.239,.334] \\
$T_k^{(1)}$ (partitioning) & $0^\dagger$ & $[2.86,7.26]\times10^{-3}$ & [0.01,0.03] & [0.02,0.04] & [0.02,0.04] & [0.03,0.06] \\
$(T_k^{(0)}-T_k^{(1)})/T_k^{(0)}$ & $100\%^\dagger$ & [94.5,97.4]\% & [82.2,93.0]\% & [84.3,90.7]\% & [85.4,91.1]\% & [80.7,89.8]\% \\
\bottomrule
\end{tabular}
\end{table}

At slot 1, the population floor is zero and the recovered fraction is $100\%$ by identity. The raw
solver residual is $\ensuremath{1\times10^{-3}}$, with bootstrap interval $[\ensuremath{6\times10^{-4}},\ensuremath{1.5\times10^{-3}}]$; we use it only as a numerical
resolution diagnostic (\cref{app:noise}). At slot 6, the $95\%$ prompt-bootstrap intervals for
$G^{(0)}/R$ are $[49.4,60.5]\%$ on Qwen3-4B, $[36.2,50.5]\%$ on Qwen3-8B,
$[38.6,53.8]\%$ on Qwen3-14B, and $[59.2,69.5]\%$ on Gemma-4-12B.

\section{Additional analyses of target continuation structure}\label{app:additional-results}

This appendix expands the continuation-concentration and locality analyses in the main paper
and examines whether path-conditioned target distributions cluster around a few recurring modes.

\subsection{Continuation concentration}\label{app:concentration}

To characterise the branching underlying the locality result in \cref{sec:q2}, we measure the
collision-equivalent effective support of the target's distribution over six-token prefixes,
$N_{\mathrm{eff}}^{(2)}=1/\sum_z p(z)^2$. It is close to one for a nearly deterministic continuation
and grows as probability spreads across multiple paths. The estimator is given in
\cref{app:effective-support}. Effective support is small for most anchors. With $M=1024$ target rollouts, its median is $1.8$ and its ninetieth percentile is $18.9$.
Both are well below the value $M$ recorded when no pair of rollouts collides.

At slot 6, effective support spans more than two orders of magnitude across the 384 anchors. Its
logarithm has a H\'ajek-weighted correlation of $+0.90$ with $T_6^{(0)}$. The correlation remains
high within each domain: $+0.845$ on alpaca, $+0.915$ on arena-hard, $+0.920$ on gsm8k, and
$+0.904$ on mbpp. On the raw scale, the corresponding correlation is $+0.53$ overall and ranges
from $+0.53$ to $+0.65$ within domains. Thus, the average floor mixes many nearly deterministic
blocks with a smaller set of highly branching ones. A complementary analysis of whether the
path-conditioned target distributions cluster around a few prototypes follows in \cref{app:trees}.

\subsection{Oracle-routed \texorpdfstring{$K$}{K}-median}\label{app:trees}

\begin{figure}
\centering
\includegraphics[width=\linewidth]{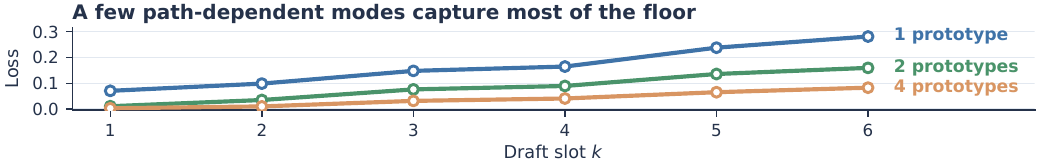}
\caption{Oracle-routed loss with $K$ fixed prototype distributions, for $K=1,2,4$.
The horizontal axis $k$ indexes draft slots.}
\label{fig:prototype-modes}
\end{figure}

The concentration analysis in \cref{app:concentration} describes diversity among continuations,
but different paths may still induce similar next-token distributions. We therefore ask whether these distributions are
widely dispersed or cluster around a few recurring modes. The $K$-median experiment fixes $K$
prototype distributions $q^{[1]},\ldots,q^{[K]}$ before the rollout and grants an oracle that assigns
each realised path to its nearest prototype. Its objective is
\[
\min_{q^{[1]},\ldots,q^{[K]}}\E_{\mu_x}\!\left[\min_{1\le j\le K}\TV(p_k,q^{[j]})\right].
\]
At $K=1$, this is exactly $T^{(0)}$. For $K>1$, we use Lloyd alternation: assign each realised path
to its nearest proposal, then replace each proposal with the TV barycentre of its assigned paths.
We run twelve updates from each of three initialisations over 128 anchors with $M=256$ paths.
\Cref{fig:prototype-modes} shows the resulting oracle-routed losses for $K=1,2,4$.

At slot 6, two prototypes remove $43\%$ of the order-0 floor and four remove $71\%$. The same
pattern holds across the block, suggesting that a small number of modes captures most of the
variation among path-conditioned target distributions. This analysis measures clusterability,
not deployable drafter performance: selecting the nearest prototype requires observing the realised
path. A concrete multi-candidate drafter would need its own routing and verification rules.

At $K=1$, the implementation directly returns the TV barycentre and reproduces $T^{(0)}$ on the same
paths. For $K>1$, we retain the lowest objective across the three initialisations. Each returned
solution is feasible, so its objective upper-bounds the global $K$-median optimum and its reported
reduction is a lower bound.

Restart spread measures sensitivity to initialisation. Under the anchor inclusion weights, $80\%$
of the cell mass has zero spread and the ninetieth percentile is $0.049$. The weighted mean is at
most $0.034$ for every $(K,\text{slot})$ pair, although a thin tail reaches $0.48$. Agreement across
restarts does not certify global optimality.

\subsection{Locality of missing path information}

\cref{tab:mi-locality} gives the slot-wise values behind the mutual-information result in
\cref{sec:q2}, measured on 698 anchors over 221 prompts. Estimator configuration and sensitivity
appear in \cref{app:method,app:mi-sensitivity}.

\begin{table}
\centering\footnotesize
\caption{Fraction of missing path information recovered by the last $m$ realised tokens on the
698-anchor eligible sample, using $M=512$ paths without an effective-sample-size gate.
Dashes mark slots where those tokens comprise the full realised prefix;
intervals are $95\%$ prompt-bootstrap confidence intervals.}
\label{tab:mi-locality}
\setlength{\tabcolsep}{5pt}
\begin{tabular}{lccc}
\toprule
slot $k$ & $\rho_{k,1}$ & $\rho_{k,2}$ & $\rho_{k,4}$ \\
\midrule
2 & $94.2\%$ [86.6, 98.5] & -- & -- \\
3 & $95.8\%$ [92.4, 98.2] & $98.8\%$ [97.2, 99.8] & -- \\
4 & $94.9\%$ [92.6, 96.8] & $99.4\%$ [98.8, 99.9] & -- \\
5 & $94.6\%$ [89.7, 97.6] & $99.0\%$ [98.1, 99.9] & $99.9\%$ [99.8, 100.0] \\
6 & $92.9\%$ [86.3, 97.3] & $98.7\%$ [97.4, 99.9] & $99.7\%$ [99.4, 99.9] \\
\midrule
pooled & $\mathbf{94.3\%}$ [91.8, 96.2] & $\mathbf{99.0\%}$ [98.3, 99.5]
& $\mathbf{99.8\%}$ [99.7, 99.9] \\
\bottomrule
\end{tabular}
\end{table}

The main estimate uses all 3490 eligible order-1 anchor--slot cells. Applying
$\mathrm{ESS}\ge32$ retains 3234 cells and gives $95.2\%$ recovery, while
$\mathrm{ESS}\ge M/2$ retains 2234 cells and gives $93.9\%$. The locality conclusion therefore
persists after excluding cells with low effective sample size (\cref{app:mi-sensitivity}).

The locality persists throughout the block: one realised token recovers $92.9$--$95.8\%$ of the
missing information at every measured slot, and two tokens recover at least $98.7\%$.

\section{Drafter decomposition details}\label{app:drafter-details}

This appendix reports the numerical decompositions behind \cref{fig:stack}. DFlash is compared
with the order-0 floor, while DSpark separates its order-1 model gap from the exposure caused by
using its own predecessor. Components are aggregated before rounding, so displayed sums can differ
slightly from the reported totals.

\subsection{DFlash: order-0 floor and model gap}

\cref{tab:dflash-details} reports the order-0 floor, observed risk, and model gap at every slot.

\begin{table}
\centering\small
\caption{DFlash rejection-risk decomposition on Qwen3-4B, pooled across four domains with $M{=}256$
and full-vocabulary evaluation. Intervals are $95\%$ prompt-bootstrap confidence intervals for $G^{(0)}$.}
\label{tab:dflash-details}
\setlength{\tabcolsep}{4pt}
\begin{tabular}{lccccccc}
\toprule
slot & 0 & 1 & 2 & 3 & 4 & 5 & 6 \\
\midrule
$\Tz$ & 0.00 & 0.08 & 0.12 & 0.17 & 0.21 & 0.25 & 0.29 \\
$R$   & 0.14 & 0.24 & 0.35 & 0.43 & 0.50 & 0.57 & 0.64 \\
$G^{(0)} = R - \Tz$ & 0.14 & 0.16 & 0.23 & 0.25 & 0.29 & 0.32 & 0.35 \\
95\% CI on $G^{(0)}$ & \scriptsize[.094,.183] & \scriptsize[.122,.203] & \scriptsize[.176,.278] & \scriptsize[.204,.307] & \scriptsize[.242,.345] & \scriptsize[.268,.378] & \scriptsize[.300,.401] \\
$G^{(0)}/R$ & 100\% & 67.3\% & 65.0\% & 59.5\% & 58.6\% & 56.8\% & 55.0\% \\
\bottomrule
\end{tabular}
\end{table}

\subsection{DSpark: order-1 model gap and exposure}

\cref{tab:o1} gives the numerical values behind the right panel of \cref{fig:stack}.

\begin{table}
\centering\small
\caption{DSpark decomposition on Qwen3-4B, pooled across four domains with $M{=}256$ and
full-vocabulary evaluation. Intervals are $95\%$ prompt-bootstrap confidence intervals.}
\label{tab:o1}
\setlength{\tabcolsep}{7pt}
\begin{tabular}{lccc}
\toprule
slot & 2 & 4 & 6 \\
\midrule
$\To$ (split-half) & \ensuremath{4.8\times10^{-3}} & 0.03 & 0.04 \\
$R^{\mathrm{oracle}}$ & 0.21 & 0.28 & 0.37 \\
$G^{(1)}$ & 0.20 & 0.26 & 0.33 \\
95\% CI on $G^{(1)}$ & [.155,.252] & [.203,.315] & [.273,.381] \\
$E^{\mathrm{expo}}$ & 0.12 & 0.18 & 0.21 \\
$R^{\mathrm{self}}$ & 0.32 & 0.46 & 0.58 \\
\bottomrule
\end{tabular}
\end{table}

The split-half floor is defined on predecessor groups covering $c_k=0.98$--$0.99$ of path mass, while
the risk rows average all paths. Let $T_{k,\mathrm{cov}}^{(1)}$ denote the reported floor and
$T_{k,\mathrm{uncov}}^{(1)}$ the floor on the remaining mass. The all-path empirical floor satisfies
\[
T_{k,\mathrm{all}}^{(1)}
=c_kT_{k,\mathrm{cov}}^{(1)}+(1-c_k)T_{k,\mathrm{uncov}}^{(1)},
\qquad 0\le T_{k,\mathrm{uncov}}^{(1)}\le1.
\]
Consequently,
\[
R_k^{\mathrm{oracle}}-c_kT_{k,\mathrm{cov}}^{(1)}-(1-c_k)
\le G_k^{(1)}
\le R_k^{\mathrm{oracle}}-c_kT_{k,\mathrm{cov}}^{(1)}.
\]
The unmatched mass can therefore change either component by at most $1-c_k\le0.02$; the displayed
rows use $T_{k,\mathrm{cov}}^{(1)}$ and form an approximate finite-sample decomposition.

\section{Serving analysis and interventions}\label{app:serving}

This appendix separates three serving questions. We first measure how survival changes the path
population and how accept factors depend across slots. We then derive frontier-scale risk bounds
from published accepted lengths. Finally, we define the single-slot oracle used to rank
interventions and study repeated coordinate updates.

\subsection{Serving-risk reweighting}

The serving risk in \eqref{eq:serving-risk} reweights a fixed drafter's free rollouts by their
probability of reaching each slot. The free and serving estimates use the same paths, so their
difference is paired within each anchor, and \eqref{eq:serving-risk} is formed as a ratio of
population sums: each anchor contributes its own arrival mass $\E[W_{k-1}]$ to the denominator, so an
anchor the serving path rarely reaches contributes proportionally less. For DSpark, the proposal at
a reached slot conditions on
the target-realised predecessor, which equals its own accepted predecessor on that event.
Components are aggregated before rounding, so differences computed from the displayed values can vary
slightly from those reported. \cref{tab:serving-risk-details} reports the Qwen3-4B results pooled across four domains
with $M=256$:

\begin{table}
\centering\scriptsize
\caption{Free-rollout and serving-weighted risk for both released drafters.}
\label{tab:serving-risk-details}
\setlength{\tabcolsep}{3.5pt}
\begin{tabular}{lcccccc}
\toprule
slot $k$ & 1 & 2 & 3 & 4 & 5 & 6 \\
\midrule
DFlash $R_k$ & 0.24 & 0.35 & 0.43 & 0.50 & 0.57 & 0.64 \\
DFlash $R_k^{\mathrm{serve}}$ & 0.17 & 0.20 & 0.17 & 0.19 & 0.22 & 0.21 \\
\quad difference & $-0.07$ & $-0.15$ & $-0.26$ & $-0.31$ & $-0.35$ & $-0.42$ \\
\quad 95\% CI & $[-.098,-.047]$ & $[-.185,-.112]$ & $[-.312,-.211]$ & $[-.366,-.254]$ & $[-.421,-.287]$ & $[-.494,-.357]$ \\
\midrule
DSpark $R_k^{\mathrm{oracle}}$ & 0.14 & 0.21 & 0.27 & 0.29 & 0.35 & 0.37 \\
DSpark $R_k^{\mathrm{serve}}$ & 0.10 & 0.15 & 0.13 & 0.14 & 0.18 & 0.16 \\
\quad difference & $-0.03$ & $-0.05$ & $-0.13$ & $-0.15$ & $-0.17$ & $-0.21$ \\
\quad 95\% CI & $[-.056,-.017]$ & $[-.081,-.033]$ & $[-.173,-.097]$ & $[-.194,-.101]$ & $[-.224,-.120]$ & $[-.262,-.159]$ \\
\bottomrule
\end{tabular}
\end{table}

Slot 0 is omitted because $W_{-1}=1$, making its free and serving risks identical. For later slots,
the reached-path law is well defined because the earlier proposals are fixed. One could minimise
risk under this policy-specific law while holding those earlier proposals fixed, but the result
would depend on the policy that created the population. If the earlier proposals were optimised as
part of the same per-slot objective, they could lower conditional risk by rejecting hard paths
before the slot. The policy-independent serving objective is therefore accepted length, a sequential
optimisation outside the per-slot floor framework.

\subsection{Dependence among accept factors}\label{app:serving-details}

To quantify dependence among accept factors, define
\[
D_k:=\frac{\E_\mu[\prod_{i\le k}a_i]}{\prod_{i\le k}\E_\mu[a_i]}.
\]
The numerator is joint survival through slot $k$, while the denominator is the survival predicted
from the marginal acceptance rates. Thus $D_k>1$ means the independence calculation understates
joint survival. \cref{tab:accept-dependence} reports this ratio across the block.

\begin{table}
\centering\small
\caption{Joint survival increasingly exceeds the product of marginal acceptance rates.}
\label{tab:accept-dependence}
\setlength{\tabcolsep}{5pt}
\begin{tabular}{lccccccc}
\toprule
slot $k$ & 0 & 1 & 2 & 3 & 4 & 5 & 6 \\
\midrule
DFlash & 1.000 & 1.093 & 1.337 & 1.945 & 3.133 & 5.692 & 12.33 \\
DSpark & 1.000 & 1.039 & 1.110 & 1.310 & 1.580 & 1.986 & 2.63 \\
\bottomrule
\end{tabular}
\end{table}

Using the joint survival terms gives $\tau=4.574$ for DFlash and $5.232$ for DSpark. Multiplying the
marginal acceptance rates gives $3.397$ and $4.386$. Both calculations use the same recorded paths;
the difference comes from retaining dependence among accept factors.

\subsection{Frontier-scale serving bounds}\label{app:frontier}

The published vLLM draft-length table uses probabilistic acceptance with thinking enabled and
temperature one; it reports accepted lengths to two decimal places \citep{vllm2026dspark}. The same
source separately reports the batch-1 coding observation at $\gamma=7$ with thinking disabled.
Let $r=\max_k R_k$. The Fr\'echet inequality gives the assumption-free survival bound
\[
\E_\mu\!\left[\prod_{i\le j}a_i\right]
\geq \left[1-\sum_{i\le j}R_i\right]_+
\geq \left[1-(j+1)r\right]_+ .
\]
Substituting this into \eqref{eq:tau} yields
\[
\tau-1 \geq \sum_{j=0}^{\gamma-1}\left[1-(j+1)r\right]_+,
\]
which can be inverted numerically to lower-bound the worst per-slot risk. For the official
DeepSeek-V4-Pro DSpark checkpoint, the reported accepted length of about $5.0$ for batch-1 coding
with thinking disabled at $\gamma=7$ gives $r\gtrsim0.107$. The thinking-enabled table gives
$2.67$ for roleplay at $\gamma=6$, implying
$r\gtrsim0.233$.

The first point in the reported draft-length sweep avoids this inversion. With one speculative
token, $\tau_{\gamma=1}=1+\E_\mu[a_0]$ identically. The reported category average
$\tau_{\gamma=1}=1.82$ therefore gives $R_0\approx0.180$ to its reporting precision. Because
$T_0^{(0)}=0$ under every trajectory law, $G_0^{(0)}=R_0\approx0.180$.
The deeper-slot bounds and the API floors use different traffic and sampling laws, so they are
compared only in scale and are not subtracted.

\subsection{Single-slot oracle objective}\label{app:bestresponse}

Let $q^{\mathrm{base}}$ be the released drafter, and let $\tau_X$ denote accepted length conditional
on prefix $X$. The slot-$k$ action has the same conditioning structure as the released drafter. For
DFlash, the information cell is a singleton and the action is one distribution $q_k(\cdot\mid X)$.
For DSpark, a cell is a realised predecessor $u=Z_{k-1}$ and the action is the conditional map
$u\mapsto q_k(\cdot\mid X,u)$. At slot 0, both action spaces contain one cell. Write $\mathcal U_k$
for the relevant cells, $q_u$ for the distribution assigned to cell $u$, and
$q_{-k}^{\mathrm{base}}$ for the released mappings at all other slots. The oracle solves
\begin{equation}
q_k^{\mathrm{BR}}(X,\cdot)
\in
\arg\max_{\{q_u:u\in\mathcal U_k\}}
\tau_X\!\left(\{q_u\}_{u\in\mathcal U_k},q_{-k}^{\mathrm{base}}\right).
\label{eq:prefix-br}
\end{equation}
Thus only slot $k$ changes; the other $\gamma-1$ proposals remain fixed at their released values.
We next reduce this objective to a separable optimisation within each information cell.

For a fixed $X$, accepted length satisfies
$\tau_X - 1 = \sum_j \E_{\mu(\cdot\mid X)}[\prod_{i \le j} a_i]$. Split the sum at $k$: terms with $j < k$ do
not contain $a_k$ and collect into a constant $\kappa_{k,X}$, while every term with $j \ge k$ contains
$a_k$ exactly once and factors as $W_{k-1}\, a_k \prod_{k < i \le j} a_i$. Summing those,
\begin{equation}
\tau_X - 1 \;=\; \kappa_{k,X} + \E_{\mu(\cdot\mid X)}\big[\,W_{k-1}\,a_k\,F_k\,\big], \qquad
F_k \;=\; 1 + \sum_{j > k} \prod_{k < i \le j} a_i ,
\label{eq:br}
\end{equation}
where $W_{k-1} = \prod_{i<k} a_i$ is the reach. Neither $W_{k-1}$ nor $F_k$ involves the slot-$k$ action.
On a finite rollout sample, let $W_{k-1}^{(r)}$ and $F_k^{(r)}$ denote their values on path $r$, set
$c_r:=W_{k-1}^{(r)}F_k^{(r)}$, let $u_r$ be the path's information cell, and set
$z_r:=Z_k^{(r)}$ and $p_r:=p_T(z_r\mid X,Z_{<k}^{(r)})$. The empirical objective is
\begin{equation*}
\max_{\{q_u:u\in\mathcal U_k\}}\ \sum_{u\in\mathcal U_k}\sum_{r:u_r=u}
c_r \min\!\big(1,\ q_u(z_r)/p_r\big).
\end{equation*}
Each $q_u$ has its own simplex constraint, so the cells decouple. Within one cell, group the paths by
realised token. The objective is $\sum_v h_v(q_u(v))$ with
$h_v(t) = \sum_{r : u_r=u,\,z_r = v} c_r \min(1, t/p_r)$, each concave and piecewise linear with
right-derivative $g_v(t) = \sum_{r:\,u_r=u,\, z_r = v,\ p_r > t} c_r/p_r$, non-increasing in $t$ with
breakpoints at the distinct $p_r$ in group $v$. A separable concave maximisation over the simplex is
solved exactly by allocating mass in decreasing order of marginal slope, stopping at each breakpoint
until the unit budget is spent. Let $g_v(t^-)$ denote the left derivative. At an optimum, the KKT
conditions require
\[
g_v(q_u(v)) \le \xi_u \le g_v(q_u(v)^-) \quad\text{when }q_u(v)>0,
\qquad g_v(0)\le\xi_u \quad\text{when }q_u(v)=0,
\]
for a common multiplier $\xi_u$ within cell $u$. The water-filling allocation satisfies these
conditions by construction and is run independently in every predecessor cell for DSpark.

Tokens never realised on any path have $g_v\equiv0$, so they receive no mass while any positive
slope remains. If the budget is not exhausted after all slopes reach zero, the objective is flat in
the remaining mass and its placement is immaterial. The implementation distributes this leftover
according to the shipped proposal, which makes the un-smoothed setting well defined.

The optimiser may choose a different action at every measured prefix and, for DSpark, in every
predecessor cell, with no requirement that one shared drafter produce all of these choices. Its
average gain therefore gives an oracle upper bound on improving slot $k$ alone within the drafter's
information class. We estimate this gain by cross-fitting below.

\subsection{Full single-slot profiles}\label{app:single-slot-results}

We estimate each prefix-specific action by two-fold cross-fitting within an anchor: fit on one half
of $M=1024$ paths, score on the other, then swap and average. The table reports held-out gains over
384 anchors; intervals use a prompt-level cluster bootstrap.

\begin{center}\small
\setlength{\tabcolsep}{4pt}
\begin{tabular}{lccccccc}
\toprule
slot $k$ & 0 & 1 & 2 & 3 & 4 & 5 & 6 \\
\midrule
DFlash $\Delta\tau_k^{\mathrm{BR}}$ & 0.221 & 0.183 & 0.166 & 0.124 & 0.135 & 0.093 & 0.056 \\
95\% CI & \scriptsize[.148,.300] & \scriptsize[.143,.229] & \scriptsize[.129,.209] & \scriptsize[.092,.161] & \scriptsize[.087,.208] & \scriptsize[.059,.136] & \scriptsize[.038,.077] \\
DSpark $\Delta\tau_k^{\mathrm{BR}}$ & 0.266 & 0.213 & 0.232 & 0.185 & 0.146 & 0.137 & 0.062 \\
\bottomrule
\end{tabular}
\end{center}

Slot 6 is the minimum in all four domains; slot 0 is the maximum in three, while mbpp peaks at slot 1.
The largest in-sample minus held-out difference is $2.3\times10^{-3}$ for DFlash and $5.2\times10^{-3}$ for DSpark, and
mixing the fitted actions toward the released proposals preserves the endpoint ordering.

\subsection{Coordinate sweeps}\label{app:iterbr}

The seven gains in \cref{app:single-slot-results} are measured separately around the released
proposal. To measure their interaction, we repeatedly refit each slot against the latest actions at
the others. We use the same fit/held-out split throughout and run both sweep directions.

\begin{center}\small
\begin{tabular}{lccc}
\toprule
 & sum of isolated gains & sweep $0\!\to\!6$ & sweep $6\!\to\!0$ \\
\midrule
DFlash & 0.98 & 2.15 \scriptsize[1.883, 2.430] & 2.07 \scriptsize[1.794, 2.357] \\
DSpark & 1.24 & 2.45 \scriptsize[2.127, 2.794] & 2.46 \scriptsize[2.128, 2.796] \\
\bottomrule
\end{tabular}
\end{center}

The sweep gain is about twice the sum of the isolated gains: $2.20\times$ for DFlash and
$1.98\times$ for DSpark. Updating a slot changes both the reach of later slots and the continuation
value of earlier ones, so subsequent updates solve different problems. The two directions agree
within $0.001$ for DSpark and differ by $0.085$ for DFlash. The sweep gives every slot a
prefix-specific action; its gain measures interaction within this oracle class and does not estimate
trainable headroom or the joint optimum.

\section{Architectural interpretation of the model gap}\label{app:interpretation}

This appendix separates the information floor from the best risk attainable within a fixed
architecture and shows why the difference can be large.

\subsection{Decomposing the model gap}\label{app:arch}

The information floor minimises risk over all proposal mappings with access to $\Iset_m$. A fixed
drafter architecture may realise only a smaller class $\mathcal{Q}_{\mathrm{arch}}$. Its best-in-class
risk is
\begin{equation}
T_{k,\mathrm{arch}}^{(m)}
:=\inf_{q\in\mathcal{Q}_{\mathrm{arch}}}
  \E_\mu\!\left[\TV\!\left(p_k,q(\cdot\mid\Iset_m)\right)\right].
\end{equation}
For a released proposal in this class,
\begin{equation}
T_k^{(m)} \le T_{k,\mathrm{arch}}^{(m)} \le R_k,
\qquad
G_k^{(m)}
=
\underbrace{R_k-T_{k,\mathrm{arch}}^{(m)}}_{\text{best-in-class gap}}
+
\underbrace{T_{k,\mathrm{arch}}^{(m)}-T_k^{(m)}}_{\text{architectural slack}}.
\label{eq:arch}
\end{equation}
The first term is the largest improvement available within the architecture class; the second is the
cost of restricting the proposal mapping to that class. Hence $G_k^{(m)}$ upper-bounds best-in-class
improvement, while realised training gains may be smaller. The information ceiling
$1-T_k^{(m)}$ remains valid because restricting the proposal class can only increase risk.

\subsection{Architectural slack can be large}

Consider a uniform input $X\in\{0,1\}^n$. Let the target's first token be $a$ when
$\bigoplus_iX_i=0$ and $b$ otherwise. The target conditional is deterministic for every $X$, so
$T_0^{(0)}=0$. For any proposal class whose outputs are independent of the parity label under the
uniform law, the expected probability assigned to the correct token is at most $1/2$. Hence
$T_{0,\mathrm{arch}}^{(0)}\ge1/2$, and the constant proposal assigning $1/2$ to each token attains
equality. Its risk and model gap are both $1/2$ although the information floor is zero.

Circuit results give related exact-computation separations. Generalised unique-hard-attention
transformers are contained in $\mathrm{AC}^0$ \citep{hao2022formal}, while parity is not in
$\mathrm{AC}^0$ \citep{furstparity,hastad1986almost}; this rules out zero risk for the construction
in that formal model. Log-precision transformers in the model of
\citet{merrill2023parallelism} are contained in logspace-uniform $\mathrm{TC}^0$, which contains
parity. An analogous separation can use an $\mathrm{NC}^1$-complete family such as the word problem
over $S_5$ \citep{barrington1986bounded}, conditional on
$\mathrm{TC}^0\ne\mathrm{NC}^1$. These results establish possible architectural slack but do not
determine its expected TV magnitude under the measured law.

The acceptance-optimal blind proposal can also be harder to compute than one target conditional. It
is the TV barycentre of the entire continuation family
$\{p_T(\cdot\mid X,Z_{<k})\}$. Computing this aggregate is a different map of $X$ from evaluating one
member of the family, so representability of each conditional does not imply representability of
the barycentre.

\section{Prefix-attention drafter: implementation and evaluation}\label{app:solution-details}

\subsection{Head parameterisation}\label{app:head-parameterisation}

Using the notation of \cref{sec:solution}, we expand $f_{\mathrm{in}}$ and $f_{\mathrm{head}}$
in \eqref{eq:prefix-head}. The input projection is
\begin{equation}
x_{k,a}=W_{\mathrm{in}}\big[\mathrm{N}_e(E(a));\mathrm{N}_h(h_k(X))\big],
\label{eq:head-query}
\end{equation}
where $\mathrm{N}$ denotes RMS normalisation and subscripts distinguish learned modules.
Separate input normalisation prevents the relative scale of the frozen embedding and drafter
state from suppressing the predecessor signal. Cross-attention and feed-forward sublayers give
\begin{align}
\widetilde{x}_{k,a}
&=x_{k,a}+W_O\,\operatorname{MHA}\!\left(
    \mathrm{N}_a(x_{k,a}),\mathrm{N}_c(H_X),\mathrm{N}_c(H_X)\right),
\label{eq:head-attention}\\
z_{k,a}
&=\widetilde{x}_{k,a}+\operatorname{MLP}\!\left(\mathrm{N}_f(\widetilde{x}_{k,a})\right),
\qquad
\ell_k(X,a)=u_k(X)+B\,\mathrm{N}_o(z_{k,a}).
\label{eq:head-readout}
\end{align}
Thus $f_{\mathrm{head}}(x_{k,a},H_X)=\mathrm{N}_o(z_{k,a})$ in the main text. The attention
projections include per-head query and key normalisation, and the MLP uses SwiGLU. Predecessor
hypotheses share projected prefix keys and values without attending to one another.

\subsection{Final architecture and training}\label{app:solution-training}

The parallel drafter has five layers and hidden width 2560. It consumes target features from layers
$[1,9,17,25,33]$. The prefix-attention head uses residual width 512, four attention heads of
dimension 128, and a SwiGLU hidden width of 2048. The predecessor embedding is borrowed from the
frozen target embedding table. Its output vocabulary map has shape $V\times512$, where
$V=151{,}936$. Removing the vanilla predecessor table makes this map's parameter count equal to
that of the vanilla rank-256 predecessor and successor tables together:
$512V=2\times256V=77{,}791{,}232$. The attention projections and MLP add parameters beyond those
tables; the entire new head is not parameter-neutral.

Prefix keys and values are projected from the fused target features and shared across hypotheses.
The committed token enters the drafter input and supplies the predecessor
at the first prediction. Separate RMS normalisations are applied to the predecessor embedding,
drafter state, and context features; the output RMS scale is initialised to $0.35$.

Each of the 20 short-convolution modules uses the two-tap, block-causal form introduced by
DFlash2 \citep{inco2026dflash2}:
\[
\operatorname{Conv}(x)_j
=\big(b_0+c_0(x_j)\big)\odot x_j
+\big(b_1+c_1(x_j)\big)\odot x_{j-1},
\]
with the preceding term zero at the input-block boundary. The base coefficients $b_0,b_1$ are
per-channel parameters, and a linear projection of $x_j$ predicts the two corrections shared
within groups of 16 channels. We initialise $b_0=1$, $b_1=0$, and both correction projections to
zero, giving an identity operator. Our wiring differs from DFlash2: we use independent pre- and
post-sublayer modules, with the latter predicting its coefficients from the sublayer output.
DFlash2 instead predicts both sides' coefficients once from the sublayer input and carries the
output-side kernel across the sublayer. The tap structure, insertion points, and parameter count
are unchanged; at width 2560 the modules add 16,486,400 parameters. The six learned masked-position
offsets add another 15,360 parameters and provide an explicit prediction-horizon signal at masked
positions, complementing RoPE's position dependence in attention.

Training data comes from Open PerfectBlend \citep{labonne2024openperfectblend}. We split the
source conversations 95/5 with seed 42 and regenerate every assistant response with
the target, Qwen3-4B (temperature 0.7, top-$p$ 0.8, top-$k$ 20, thinking disabled), giving
1,339,867 target-generated training examples. Training uses maximum sequence length 4096 and 512
anchors per training sequence. Target features and conditional probabilities
are computed online. The target embedding and language-model output tables remain frozen.
We use seed 42, global batch size 512, learning rate $6\times10^{-4}$, warmup ratio $0.04$,
cosine decay over 26,160 steps, gradient clipping at 1.0, no weight decay, and bfloat16 precision.
We retain DSpark's cross-entropy, distributional $L_1$ and confidence
objectives, with weights $0.1$, $0.9$ and $1.0$, respectively. To train the base logits for candidate
coverage, we add a candidate-nomination loss with weight $0.1$:
\begin{equation}
\mathcal L_{\mathrm{nom},k}
=-\sum_{v\in S_k}p_k^{\mathrm{nom}}(v)\log\operatorname{softmax}(u_k)_v,
\label{eq:nomination-loss}
\end{equation}
where $S_k$ contains the target's top 16 tokens and $p_k^{\mathrm{nom}}$ is the target distribution
renormalised on $S_k$. The softmax spans the full vocabulary. All terms use slot weights
$\exp(-k/4)$.

\textbf{Inference proposal.} Candidates are nominated from the drafter's base logits.
The attention head scores each
possible predecessor against those successor candidates; conditional scores are normalised over
the nominated successors before sampling. The verifier uses that proposal in its acceptance and
residual-sampling computations.

\subsection{Serving protocol and statistical units}\label{app:solution-eval}

Within task $d$, let $n_r^{\mathrm{completion}}$ and $n_r^{\mathrm{verify}}$ be the returned
completion-token and verification-step counts for request row $r$. We report
\begin{equation}
\widehat\tau_d=
\frac{\sum_{r\in d}n_r^{\mathrm{completion}}}{\sum_{r\in d}n_r^{\mathrm{verify}}},
\qquad
\widehat\tau_{\mathrm{macro}}=\frac{1}{9}\sum_d\widehat\tau_d.
\label{eq:solution-aggregation}
\end{equation}
Returned tokens include the verifier's bonus token. This weights verification steps within each
task and differs from averaging per-request ratios.

We follow DSpark's evaluation protocol \citep{cheng2026dsparkconfidencescheduledspeculativedecoding},
using identical benchmark rows and sampling settings for both checkpoints. Both use seven
speculative positions and chain verification without confidence-based early stopping;
per-task sample counts appear in \cref{tab:solution-serving-details}.

The speedups in \cref{tab:solution-serving} use sequential requests on one H100 80GB in
bfloat16, including prefill and streaming overhead. For accepted length, we average
$\widehat\tau_{d,C}$ equally over $C\in\{1,4,8,16,32\}$ within each task, then equally
over the nine tasks. These concurrency settings use the same prompt set.

\begin{table}[ht]
\centering\small
\caption{Per-task accepted length ($\tau$) and sequential speedup ($S$) for
\cref{tab:solution-serving}. Macro averages the nine tasks equally.}
\label{tab:solution-serving-details}
\setlength{\tabcolsep}{2.5pt}
\begin{tabular}{lr*{8}{r}}
\toprule
 & & \multicolumn{4}{c}{Temperature 0} & \multicolumn{4}{c}{Temperature 1} \\
\cmidrule(lr){3-6}\cmidrule(lr){7-10}
 & & \multicolumn{2}{c}{$\tau$} & \multicolumn{2}{c}{$S$}
 & \multicolumn{2}{c}{$\tau$} & \multicolumn{2}{c}{$S$} \\
\cmidrule(lr){3-4}\cmidrule(lr){5-6}\cmidrule(lr){7-8}\cmidrule(lr){9-10}
Task & Rows & DSpark & Ours & DSpark & Ours & DSpark & Ours & DSpark & Ours \\
\midrule
GSM8K & 500 & 6.365 & \textbf{6.523} & 4.29 & \textbf{4.44} & 6.091 & \textbf{6.279} & 4.08 & \textbf{4.30} \\
MATH-500 & 500 & 6.302 & \textbf{6.451} & 4.44 & \textbf{4.59} & 5.736 & \textbf{5.893} & 4.08 & \textbf{4.21} \\
AIME25 & 30 & 5.783 & \textbf{5.949} & 4.18 & \textbf{4.35} & 4.908 & \textbf{5.127} & 3.54 & \textbf{3.78} \\
HumanEval & 164 & 5.688 & \textbf{5.829} & 3.96 & \textbf{4.10} & 5.468 & \textbf{5.677} & 3.79 & \textbf{4.04} \\
MBPP & 256 & 5.381 & \textbf{5.522} & 3.66 & \textbf{3.82} & 5.139 & \textbf{5.322} & 3.56 & \textbf{3.69} \\
LiveCodeBench & 500 & 5.763 & \textbf{5.936} & 4.01 & \textbf{4.16} & 4.937 & \textbf{5.181} & 3.47 & \textbf{3.69} \\
MT-Bench & 80 & 3.816 & \textbf{3.968} & 2.73 & \textbf{2.85} & 3.658 & \textbf{3.812} & 2.64 & \textbf{2.76} \\
Alpaca & 500 & 3.692 & \textbf{3.825} & 2.59 & \textbf{2.71} & 3.549 & \textbf{3.721} & 2.54 & \textbf{2.69} \\
Arena-Hard-v2 & 500 & 3.732 & \textbf{3.861} & 2.67 & \textbf{2.80} & 3.292 & \textbf{3.447} & 2.35 & \textbf{2.49} \\
\midrule
Macro & 3030 & 5.17 & \textbf{5.32} & 3.62 & \textbf{3.76} & 4.75 & \textbf{4.94} & 3.34 & \textbf{3.52} \\
\bottomrule
\end{tabular}
\end{table}

\begin{table}[ht]
\centering
\caption{Full-model training trajectory at temperature 1, measured before serving optimizations. Every
checkpoint belongs to the same 10-epoch schedule.}
\label{tab:solution-trajectory}
\setlength{\tabcolsep}{4pt}
\begin{tabular}{rrrrrrrrrr}
\toprule
Epoch 1 & 2 & 3 & 4 & 5 & 6 & 7 & 8 & 9 & 10 \\
\midrule
4.29 & 4.56 & 4.70 & 4.78 & 4.86 & 4.90 & 4.92 & 4.92 & 4.94 & 4.94 \\
\bottomrule
\end{tabular}
\end{table}

The trajectory in \cref{tab:solution-trajectory} belongs to one training run. Gains become smaller
late in training, with epochs seven, nine, and ten giving $4.92$, $4.94$, and $4.94$.

\subsection{Speedup against concurrency}\label{app:solution-batch}

We evaluate AR, DSpark and ours on the same 3,030 rows at both temperatures and
$C\in\{1,4,8,16,32\}$ concurrent requests. We measure throughput over each
complete client-side task run, including prefill and streaming, then average the
nine task-level ratios for each comparison.
\begin{table}[ht]
\centering\small
\caption{Serving sweep with exact end-to-end timing, including prefill and HTTP.
Each speedup is an equal mean of nine task-level ratios. $C$ denotes request concurrency.}
\label{tab:serving-sweep}
\setlength{\tabcolsep}{5pt}
\begin{tabular}{rrrrrrr}
\toprule
$T$ & $C$ & $S$ DSpark & $S$ Ours & Ours/DSpark gain & $\tau$ DSpark & $\tau$ Ours \\
\midrule
0 & 1 & 3.616 & 3.757 & $+3.99\%$ & 5.178 & 5.333 \\
0 & 4 & 3.415 & 3.536 & $+3.65\%$ & 5.169 & 5.330 \\
0 & 8 & 3.141 & 3.301 & $+5.10\%$ & 5.157 & 5.312 \\
0 & 16 & 2.896 & 3.032 & $+4.77\%$ & 5.159 & 5.308 \\
0 & 32 & 3.112 & 3.209 & $+3.35\%$ & 5.182 & 5.309 \\
\midrule
1 & 1 & 3.339 & 3.518 & $+5.40\%$ & 4.748 & 4.941 \\
1 & 4 & 3.133 & 3.222 & $+3.08\%$ & 4.761 & 4.937 \\
1 & 8 & 2.977 & 3.098 & $+4.38\%$ & 4.756 & 4.932 \\
1 & 16 & 2.689 & 2.838 & $+5.95\%$ & 4.743 & 4.933 \\
1 & 32 & 2.695 & 2.831 & $+5.27\%$ & 4.758 & 4.956 \\
\bottomrule
\end{tabular}
\end{table}

The gain ranges from 3.35--5.10\% at temperature 0 and 3.08--5.95\% at temperature 1.

\subsection{Per-slot model gap and exposure}\label{app:solution-gap}

We evaluate the final drafter on the same 384 anchors and with the same target-sampling protocol
as the DSpark decomposition in \cref{sec:q3}, using 256 paths per anchor.
\Cref{tab:solution-gap} reports the floors and risks underlying \cref{fig:solution-gap}.
The model gap decreases at slots 1--6, while slot 0 remains essentially unchanged.
At slot 6, the gap falls by $25.5\%$, from $0.33$ to
$0.24$, while exposure rises from $0.21$ to $0.26$. The self-conditioned risk therefore
decreases less, from $0.58$ to $0.54$.

\begin{table}[ht]
\centering\small
\caption{Per-slot information floor, oracle risk, and self-conditioned risk for DSpark and
the final drafter. Floors use the split-half estimate; at slot 0, $T_0^{(0)}=0$.}
\label{tab:solution-gap}
\setlength{\tabcolsep}{3pt}
\begin{tabular}{llrrrrrrr}
\toprule
Metric & Drafter & 0 & 1 & 2 & 3 & 4 & 5 & 6 \\
\midrule
$T^{(1)}$ & DSpark & 0.00 & \ensuremath{1\times10^{-3}} & \ensuremath{4.8\times10^{-3}} & 0.02 & 0.03 & 0.03 & 0.04 \\
 & Ours & 0.00 & \ensuremath{1\times10^{-3}} & \ensuremath{4.7\times10^{-3}} & 0.02 & 0.03 & 0.03 & 0.04 \\
\addlinespace[2pt]
$R^{\mathrm{oracle}}$ & DSpark & 0.11 & 0.14 & 0.21 & 0.27 & 0.28 & 0.35 & 0.37 \\
 & Ours & 0.11 & 0.11 & 0.14 & 0.19 & 0.19 & 0.27 & 0.28 \\
\addlinespace[2pt]
$R^{\mathrm{self}}$ & DSpark & 0.11 & 0.22 & 0.32 & 0.40 & 0.46 & 0.54 & 0.58 \\
 & Ours & 0.11 & 0.21 & 0.29 & 0.38 & 0.43 & 0.51 & 0.54 \\
\bottomrule
\end{tabular}
\end{table}

\subsection{Scope of the one-epoch component comparisons}\label{app:solution-variants}

The variants in \cref{tab:solution-components} were trained for 2,616 steps on a one-epoch
schedule and evaluated on the same 430 temperature-zero rows, with a per-task cap of 50 and
maximum generation length 2048. Within-task ratios follow the main table. The one-epoch
checkpoints are released at \url{https://huggingface.co/TIE-Pilot/deepspec-drafter-ablations}.

\begin{table}[ht]
\centering
\caption{One-epoch design comparisons on a separate 430-row evaluation. The DFlash2 reproduction
uses the Speculators recipe \citep{speculators2025}; other convolution rows use our wiring above.}
\label{tab:solution-components}
\begin{tabular}{lrr}
\toprule
One-epoch configuration & Accepted length & Change from vanilla \\
\midrule
Vanilla DSpark & 4.56 & --- \\
DFlash2 reproduction & 4.44 & $-0.12$ \\
Vanilla + our short convolution & 4.61 & $+0.05$ \\
Vanilla + slot embeddings & 4.56 & $-0.01$ \\
Prefix-attention head & 4.68 & $+0.11$ \\
Head + our convolution + slot embeddings & 4.77 & $+0.20$ \\
\bottomrule
\end{tabular}
\end{table}

At this budget, reproduced DFlash2 is below vanilla; the head and short-convolution arms improve
acceptance, and the combined variant is best. Slot embeddings alone are essentially unchanged,
but we retain them because they add no inference cost and appeared to accelerate convergence.
These comparisons do not isolate final component contributions, for the reasons below.

The short-convolution and slot-embedding arms each add the named component to the internal
vanilla DSpark baseline. The head variant changes the head architecture and adds the nomination
objective. Its initial output scale is approximately $4.26\times10^{-2}$; the combined variant uses 0.35
and adds short convolution and slot embeddings. Both share the vanilla baseline's backbone
initialisation.

Both the head and combined variants include a separate anchor key/value column. Removing this
column at evaluation did not affect acceptance, while retaining it caused longer runs to collapse,
so we omit it. The head variant's output-dependent gate, intended to approximate ordinary
self-attention to the current token, also showed no benefit and is omitted from the combined and
final configurations.

\end{document}